%% file: main.tex
\documentclass[runningheads]{llncs}

\newcommand{\ieee}[1]{}
\newcommand{\lncs}[1]{#1}
\newcommand{\acm}[1]{}
\newcommand{\blind}[1]{#1}

\input{Premeables/packages}

\input{Premeables/defs}
\input{Premeables/acronyms}

\begin{document}
\ieee
{
    \title{
Learning a Behavior Model of Hybrid Systems Through Combining Model-Based Testing and Machine Learning
  
        \blind
        {
            \thanks{This work is supported by the Graz University of Technology's LEAD project
            ``Dependable Internet of Things in Adverse Environments''. 
            It is also partially supported by the ECSEL Joint Undertaking (ENABLE-S3, grant 692455).}
        }
    }
}

\lncs{
\begin{full}
\title{Learning a Behavior Model of Hybrid Systems Through Combining Model-Based Testing and Machine Learning
\\ (Full Version)\thanks{This work is an extended preprint of the conference paper ``Learning a Behavior Model of Hybrid Systems Through Combining Model-Based Testing and Machine Learning'' accepted for presentation at IFIP-ICTSS 2019, the 
31\textsuperscript{st} International Conference on Testing Software and Systems
 in Paris, France.  
}
}
\end{full}
\begin{conference}
\title{
Learning a Behavior Model of Hybrid Systems Through Combining Model-Based Testing and Machine Learning
}
\end{conference}
}
\acm
{
    \subtitle
    {
        A Platooning Case Study
    }
    \remove
    {
        \subtitlenote
        {
            Produces the permission block, and
            copyright information
        }
        \titlenote
        {   
        }
    }
}  
\lncs
{
    \titlerunning{Learning a Behavior Model of Hybrid Systems}
}
\blind
{
    \input{Premeables/authors}

}
\acm
{
    \input{Premeables/ccs}
    \input{Sections/abstract}
}
\maketitle
\ieee
{
    \input{Sections/abstract}
}
\lncs
{
    \input{Sections/abstract}
}
\input{Sections/introduction}
\input{Sections/preliminaries}
\input{Sections/method}

\begin{full}
 \input{Sections/method_rnn_full}
\end{full}
\begin{conference}
 \input{Sections/method_rnn_conf}
\end{conference}
\input{Sections/results}
\input{Sections/related}

\input{Sections/conclusion}
\paragraph*{Acknowledgment.}
This work is supported by the TU Graz LEAD project ``Dependable Internet of Things in Adverse Environments''. It is also partially supported by 
ECSEL Joint Undertaking under Grant No.: 692455.
\lncs
{
\renewcommand{\doi}[1]{doi: \url{#1}}
\renewcommand{\doi}[1]{}
\bibliographystyle{splncs04}
\bibliography{main}
}
\ieee
{
\bibliographystyle{IEEEtranN}
\bibliography{IEEEabrv,main}
}

\acm
{
\bibliographystyle{unsrtnat}
\bibliographystyle{ACM-Reference-Format}
\bibliography{main}
}
\end{document}

%% file: Premeables/packages.tex
\usepackage{graphicx}

\usepackage{hyperref}

\ieee
{
    \IEEEoverridecommandlockouts
    \usepackage[numbers,sort]{natbib}
    
    \usepackage{amsthm}
}
\lncs
{
    \usepackage{cite}
}
\acm
{
    \usepackage{amsthm}
}

\usepackage{amssymb, amsmath, amsfonts}
\usepackage{nicefrac}
\usepackage{mathalfa}
\usepackage{flushend}
\usepackage{dsfont}
\usepackage{xspace}
\usepackage{wrapfig}
\usepackage{marvosym}
\usepackage{tikz}
\usetikzlibrary{arrows,shapes,automata,positioning,calc,shapes.symbols,shapes.geometric,shadows}
\usetikzlibrary{decorations.pathmorphing,decorations.pathreplacing,decorations.shapes,chains,fit}
\usepackage{pgfplots, pgfplotstable}
\pgfplotsset{compat=1.14}
\usepackage{rotating}
\usepackage{pifont}
\usepackage{multirow}
\usepackage{multicol}
\usepackage[utf8]{inputenc}
\usepackage[style=base]{caption}
\usepackage{acronym}
\usepackage{subcaption}
\usepackage{array}
\usepackage{booktabs}
\usepackage{makecell}%
\usepackage[noend]{algpseudocode}
\usepackage{algorithm}
\usepackage[capitalise]{cleveref}
\usepackage{enumitem}%
\usepackage{makecell}%
\usepackage{xcolor}%
\usepackage[noend]{algpseudocode}
\usepackage{xargs}                      
\usepackage
[colorinlistoftodos,
prependcaption,
textsize=small,
textwidth=6cm
]{todonotes}

%% file: Premeables/defs.tex
\newcommand\iaik{Institute of Applied Information Processing and Communications}
\newcommand\spsc{Signal Processing and Speech Communication Laboratory}
\newcommand\ist{Institute of Software Technology}
\newcommand\irt{Institute of Automation and Control}
\newcommand\tugraz{Graz University of Technology}

\newcommand\remove[1]{}

\algrenewcommand\algorithmicindent{0.50em}%
\algnewcommand\algorithmicswitch{\textbf{switch}}
\algnewcommand\algorithmiccase{\textbf{case}}
\algdef{SE}[SWITCH]{Switch}{EndSwitch}[1]{\algorithmicswitch\ #1\ \algorithmicdo}{\algorithmicend\ \algorithmicswitch}%
\algdef{SE}[CASE]{Case}{EndCase}[1]{\algorithmiccase\ #1}{\algorithmicend\ \algorithmiccase}%
\algdef{SE}[DOWHILE]{Do}{doWhile}{\algorithmicdo}[1]{\algorithmicwhile\ #1}%
\algtext*{EndSwitch}%
\algtext*{EndCase}%
\algnewcommand{\algorithmicgoto}{\textbf{go to}}%
\algnewcommand{\Goto}{\algorithmicgoto\xspace}%
\algnewcommand{\Label}{\State\unskip}

\makeatletter
\renewcommand{\ALG@beginalgorithmic}{\scriptsize}
\makeatother
\algrenewcommand\alglinenumber[1]{\scriptsize #1:}

\newcommand{\lto}[1]{\xrightarrow{#1}}
\newcommand{\sto}[1]{\xrightarrow{#1}\!\!^*}

\newcommand{\eg}{e.\,g.,\ }
\newcommand{\ie}{i.\,e.,\ }

\newcommand{\st}{s.t.\ }

\newcommand{\tool}[1]{\textsc{#1}}
\newcommand\Lstar{L${}^*$\xspace}

\newcommandx{\proj}[2][1=\mathop{\AI}]{\mathop{\left(#2 \downharpoonright #1 \right)}}

\newcommand\I{\mathrm{I}}
\renewcommand\O{\mathrm{O}}
\newcommand\X{\mathcal{X}}

\renewcommand\H{\mathcal{H}}

\newcommand\M{\mathcal{M}}
\newcommand\Init{\mathbf{Init}}
\newcommand\Inv{\mathbf{Inv}}
\newcommand\Flow{\mathbf{Flow}}
\newcommand\Jump{\mathbf{Jump}}
\newcommand\Violation{\mathit{Violations}}
\newcommand\AI{I}
\newcommand\ai{i}
\newcommand\AO{O}

\newcommand\conc{\gamma}
\newcommand\abs{\alpha}
\newcommand\ao{o}

\newcommand{\acc}{\textit{acc}\xspace}
\newcommand{\dist}{\textit{d}\xspace}
\newcommand{\vleader}{\textit{v}_\textit{l}\xspace}
\newcommand{\vfollow}{\textit{v}_\textit{f}\xspace}
\newcommand{\orientation}{\Delta\xspace}

\newcommand\wait{\textit{const}_l}
\newcommand{\crash}{\textit{crash}}
\newcommand\Ntrain{N_\mathrm{train}}
\newcommand\Nval{N_\mathrm{val}}
\newcommand\Nautlearn{N_\mathrm{autl}}

\newcommandx\bka[2][1=]
{%
    \ifthenelse{\equal{#1}{c}}
    { \textcolor{black!30!blue}{[* {\bf BKA:} #2 *]} }
    {   
        \ifthenelse{\equal{#1}{h}}
        {}
        {#2}
    }
}%

\newcommandx\mt[2][1=]
{
    \ifthenelse{\equal{#1}{c}}
    { \textcolor{black!50!green}{[* {\bf MT:} #2 *]} }
    {   
        \ifthenelse{\equal{#1}{h}}
        {}
        {#2}
    }
}%

\newcommandx{\me}[2][1=]
{
    \ifthenelse{\equal{#1}{c}}
    { \textcolor{blue!30!purple}{[* {\bf ME:} #2 *]} }
    {   
        \ifthenelse{\equal{#1}{h}}
        {}
        {#2}
    }
}%

\newcommandx\rb[2][1=]
{
    \ifthenelse{\equal{#1}{c}}
    { \textcolor{black!30!red}{[* {\bf RB:} #2 *]}   }
    {   
        \ifthenelse{\equal{#1}{h}}
        {}
        {#2}
    }
}%

\newcommandx\ar[2][1=]{
    \ifthenelse{\equal{#1}{c}}
    { \textcolor{black!30!orange}{[* {\bf AR:} #2 *]}  }
    {   
        \ifthenelse{\equal{#1}{h}}
        {}
        {#2}
    }
}%

\acm
{

\newtheoremstyle{sig}
  {}
  {}
  {\itshape}
  {}
  {\scshape}
  {.}
  {.5em}
  {#1 #2\thmnote{~(#3)}}
\theoremstyle{sig}

\newtheorem{theorem}{Theorem}[section]
\newtheorem{proposition}{Proposition}
\newtheorem{corollary}{Corollary}
\newtheorem{lemma}{Lemma}
\newtheorem{definition}{Definition}
\newtheorem{remark}{Remark}
\newtheorem{example}{Example}

}

\ieee{

\newtheorem{definition}{Definition}

}

\newcommand{\motExample}{{\textbf{Back to Motivating Example.}}\xspace}

\reversemarginpar
\newcommandx{\unsure}[2][1=]
{
\todo[linecolor=red,backgroundcolor=red!25,bordercolor=red,#1]
{#2}\xspace
}
\newcommandx{\change}[2][1=]{
\todo[linecolor=blue,backgroundcolor=blue!25,bordercolor=blue,#1]
{#2}\xspace
}
\newcommandx{\info}[2][1=]
{
\todo[linecolor=green,backgroundcolor=green!35,bordercolor=green,#1]
{#2}\xspace
}
\newcommandx{\improvement}[2][1=]
{
\todo[linecolor=Plum,backgroundcolor=Plum!25,bordercolor=Plum,#1]
{#2}\xspace
}
\newcommandx{\thiswillnotshow}[2][1=]{\todo[disable,#1]{#2}\xspace}

\input{Premeables/tikz}
\newcommand{\cdfheight}{5}
\newcommand{\plotheight}{6}
\usepackage{environ}
\usepackage{ifthen}
\newboolean{full}           
\setboolean{full}{true}

\ifthenelse{\boolean{full}}%
  {%
  \NewEnviron{full}
    {%
    \BODY{}
        }%
}%
    {\NewEnviron{full}
    {%
    }%
}%

\ifthenelse{\boolean{full}}%
  {%
  \NewEnviron{conference}
    {%
        }%
}%
    {\NewEnviron{conference}
    {%
    \BODY{}
    }%
}%

\acm
{
    \setcopyright{rightsretained}
    
    \acmConference[ISSTA 2019]
    {ACM SIGSOFT International Symposium on Software Testing and Analysis}
    {15–19 July, 2019}
    {Beijing, China}
}

%% file: Premeables/tikz.tex
\tikzstyle {line} = [draw, -latex']
\tikzstyle {box} = [ 
    rectangle, draw, 
    align=center, 
    minimum width=10mm, 
    minimum height=10mm
]

\tikzstyle {port} = [ 
    rectangle, align=center, 
    minimum width=2mm, minimum height=2mm 
]

\tikzstyle{whitebox} = [
    draw, 
    rectangle, 
    text width=1.75cm,
    fill=white, 
    text badly centered, 
    minimum height=1cm, 
    minimum width=1.6180cm
]

\tikzstyle{blackbox} = [
    whitebox,
    fill=black, 
    text=white
]

\tikzstyle{files} = [
    shape = tape,
    draw,
    fill = white,
    tape bend top = none,
    double copy shadow, 
    minimum height=1cm, 
    minimum width=1.6180cm
]

\makeatletter
\tikzset{
    database/.style={
        path picture={
            \draw (0, 1.5*\database@segmentheight) circle [x radius=\database@radius,y radius=\database@aspectratio*\database@radius];
            \draw (-\database@radius, 0.5*\database@segmentheight) arc [start angle=180,end angle=360,x radius=\database@radius, y radius=\database@aspectratio*\database@radius];
            \draw (-\database@radius,-0.5*\database@segmentheight) arc [start angle=180,end angle=360,x radius=\database@radius, y radius=\database@aspectratio*\database@radius];
            \draw (-\database@radius,1.5*\database@segmentheight) -- ++(0,-3*\database@segmentheight) arc [start angle=180,end angle=360,x radius=\database@radius, y radius=\database@aspectratio*\database@radius] -- ++(0,3*\database@segmentheight);
        },
        minimum width=2*\database@radius + \pgflinewidth,
        minimum height=3*\database@segmentheight + 2*\database@aspectratio*\database@radius + \pgflinewidth,
    },
    database segment height/.store in=\database@segmentheight,
    database radius/.store in=\database@radius,
    database aspect ratio/.store in=\database@aspectratio,
    database segment height=0.1cm,
    database radius=0.25cm,
    database aspect ratio=0.35,
}
\makeatother

\pgfplotsset{
    legend entry/.initial=,
    every axis plot post/.code={%
        \pgfkeysgetvalue{/pgfplots/legend entry}\tempValue
        \ifx\tempValue\empty
            \pgfkeysalso{/pgfplots/forget plot}%
        \else
            \expandafter\addlegendentry\expandafter{\tempValue}%
        \fi
    },
}

\makeatletter
\long\def\ifnodedefined#1#2#3{%
    \@ifundefined{pgf@sh@ns@#1}{#3}{#2}%
}

\pgfplotsset{
    discontinuous/.style={
    scatter,
    scatter/@pre marker code/.code={
        \ifnodedefined{marker}{
            \pgfpointdiff{\pgfpointanchor{marker}{center}}%
             {\pgfpoint{0}{0}}%
             \ifdim\pgf@y>0pt
                \draw [densely dashed] (marker-|0,0) -- (0,0);
                \draw[fill=white] plot [mark=*] coordinates {(marker-|0,0)};
             \else
                \tikzset{options/.style={mark=none}}
             \fi
        }{
            \tikzset{options/.style={mark=none}}
        }
        \coordinate (marker) at (0,0);
        \begin{scope}
    },
    scatter/@post marker code/.code={\end{scope}}
    }
}
\makeatother

\pgfplotsset{range/.style 2 args={
    x filter/.code={
        \ifnum\coordindex<#1\def\pgfmathresult{}\fi
        \ifnum\coordindex>#2\def\pgfmathresult{}\fi
    }
}}

%% file: Premeables/acronyms.tex
\newacro{DFA}{deterministic finite automaton}
\acrodefplural{DFA}{deterministic finite automata}

\newacro{SUL}{system under learning}
\acrodefplural{SUL}{systems under learning}

\newacro{SUT}{system under test}
\acrodefplural{SUT}{systems under test}

\newacro{LTL}{Linear Temporal Logic}
\newacro{LBT}{Learning-Based Testing}
\newacro{MBT}{Model-Based Testing}
\newacro{TCBT}{Transition-Coverage Based Testing}
\newacro{SIL}{Software-In-the-Loop}
\newacro{DT}{Decision Tree}
\newacro{LLM}{Logic Learning Machine}
\newacro{FPR}{False Positive Rate}
\newacro{FNR}{False Negative Rate}
\newacro{V2V}{Vehicle-to-Vehicle}
\newacro{MAT}{minimally adequate teacher}
\newacro{CPS}{Cyber Physical System}
\newacro{CO-CPS}{Cooperative Open Cyber Physical System}
\newacro{IoT}{Internet of Things}

\newacro{CDF}{cumulative distribution function}
\newacro{DT}{Decision Tree}
\newacro{NN}{Neural Network}
\newacro{MBDG}{Model-Based Dataset Generation}

\newacro{HA}{Hybrid Automaton}
\acrodefplural{HA}{Hybrid Automata}

\newacro{RNN}{Recurrent Neural Network}

%% file: Premeables/authors.tex
\lncs
{
    \author
    {
        Bernhard K. Aichernig \and 
        Roderick Bloem \and 
        Masoud Ebrahimi \and
        Martin Horn \and
        Franz Pernkopf \and 
        Wolfgang Roth \and
        Astrid Rupp \and
        Martin Tappler \and 
        \\Markus Tranninger 
    }
    \authorrunning{B.\,K. Aichernig et al.}
    \institute
    {
        Graz University of Technology, Graz, Austria\\
        \email{aichernig@ist.tugraz.at},  \email{roderick.bloem@iaik.tugraz.at}, \email{masoud.ebrahimi@iaik.tugraz.at}, \email{martin.horn@tugraz.at}, \email{pernkopf@tugraz.at}, \email{roth@tugraz.at}, \email{astrid.rupp@fprimezero.com}, \email{martin.tappler@ist.tugraz.at},
        \email{markus.tranninger@tugraz.at}
    }
}

%% file: Premeables/ccs.tex
%
%

%% file: Sections/abstract.tex
\begin{abstract}
{
Models play an essential role in the design process of cyber-physical systems. They form the basis for simulation and analysis and help in identifying design problems as early as possible. However, the construction of models that comprise physical and digital behavior is challenging. Therefore, there is considerable interest in learning such hybrid behavior by means of machine learning which requires sufficient and representative training data covering the behavior of the physical system adequately. In this work, we exploit a combination of automata learning and model-based testing to generate sufficient training data fully automatically.

Experimental results on a platooning scenario show that recurrent neural networks learned with this data achieved significantly better results compared to models learned from randomly generated data. In particular, the classification error for crash detection is reduced by a factor of five and a similar F1-score is obtained with up to three orders of magnitude fewer training samples.
}

\remove
{
Models play an essential role in the design process of cyber-physical systems. 
They form the basis for simulation and analysis and help in identifying design problems as early as possible.
However, the construction of models that comprise physical and digital behaviour is challenging. 
Therefore, there is considerable interest in learning such hybrid behaviour with the help of machine learning methods. 
The problem is that machine learning requires representative training data that covers the large state-space of such systems adequately. 
In this work we demonstrate the application of machine learning methods in learning a behavior model of hybrid systems. 
To this end, we used a combination of automata learning and model-based testing to a training dataset that better represents the targeted behavior of a hybrid system.
Our experiments of learning a crash-detector in the forms of a Recurrent Neural Network from a distributed controller for the platooning of vehicles show that we could reduce the classification error by a factor of five.  
}


%
\lncs
{
    \keywords
    {
        Hybrid Systems \and
        Behavior Modeling \and
        Automata Learning \and
        \acl{MBT} \and
        Machine Learning \and
        Autonomous Vehicle \and
        Platooning
    }
}
\ieee
{
    \begin{IEEEkeywords} 
        Testing,
        Verification,
        Automata Learning,
        Control Theory,
        Autonomous Vehicle,
        Platooning
    \end{IEEEkeywords}
}
\end{abstract}

\acm
{
\remove
{
    \ccsdesc[500]{Computer systems organization~Embedded systems}
    \ccsdesc[300]{Computer systems organization~Redundancy}
    \ccsdesc{Computer systems organization~Robotics}
    \ccsdesc[100]{Networks~Network reliability}
}
    
    \keywords
    {
        Hybrid Systems,
        Behavior Modeling,
        Automata Learning,
        \acl{MBT},
        Machine Learning,
        Autonomous Vehicle,
        Platooning
    }
}

%% file: Sections/introduction.tex
\section{Introduction}
\remove
{
The emergence of \ac{IoT} has made \acp{CPS} to dominate safety-critical areas.
Most recently, we are witnessing an ever-increasing reliance on \acp{CPS} that demands strict safety assurances.
To achieve safety assurances for a \ac{CPS}, one could formally specify the safety property of interest, and finally apply a model-based verification method to check the safety property should it be possible to model the \ac{CPS} formally.
Of note, formal modeling is a labor intensive and error-prone process whose difficulty level strictly depends on the complexity of the system.
Manual modeling is among the reasons formal verification methods becomes prone to human errors.
}
In \acp{CPS}, embedded computers and networks control
physical processes. Most often, \acp{CPS} interact with their surroundings based on the context and the (history of) external events through an analog interface. We use the term hybrid system to refer to such reactive systems that intermix discrete and continuous components~\cite{MannaP92}.
Since hybrid systems are dominating safety-critical areas, safety assurances are of utmost importance. However, we know that most verification problems for hybrid systems are undecidable~\cite{Henzinger:HA}.

Therefore, models and model-based simulation play an essential role in the design process of such systems.
They help in identifying design problems as early as possible and facilitate integration testing with  model-in-the-loop techniques. 
However, the construction of hybrid models that comprise physical and digital behavior is challenging. Modeling such systems with reasonable fidelity requires  expertise in several areas, including control engineering, software engineering and sensor networks \cite{Derler&12}.  

Therefore, we see a growing interest in learning such
cyber-physical behavior with the help of machine learning. Examples include helicopter dynamics \cite{Punjani&15}, the physical layer of communication protocols \cite{OShea&17}, standard continuous control problems \cite{Duan&16}, and industrial process control \cite{Spielberg&17}.  

However, in general, machine learning requires a large and representative set of training data. Moreover, for the simulation of safety-critical features, rare side-conditions need to be sufficiently covered. Given the large state-space of hybrid systems, it is difficult to gather a good training set that captures all critical behavior. Neither nominal samples from operation nor randomly generated data will be sufficient. Here, advanced test-case generation methods can help to derive a well-designed training set with adequate coverage.

\setlength{\intextsep}{0pt}%
\begin{wrapfigure}[16]{t}{.44\textwidth}
  \centering
    \input{Figures/workflow}
  \caption{Learning a behavior model of a black-box hybrid system.}\label{fig:flow}
  \end{wrapfigure}
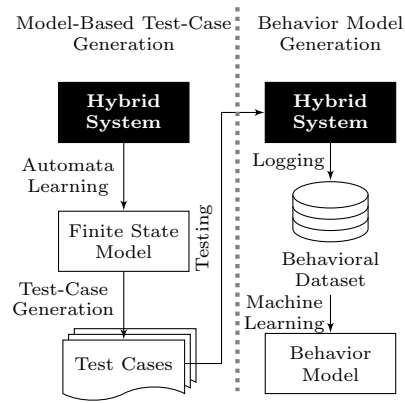
%
In this paper, we combine automata learning and \ac{MBT} to derive an adequate training set, and then use machine learning to learn a behavior model from a black-box hybrid system. 
We can use the learned behavior model for multiple purposes such as monitoring runtime behavior. Furthermore, it could be used as a surrogate of a complex and heavy-weight simulation model to efficiently analyze safety-critical behavior offline~\cite{simulation_model_approx}. 
\Cref{fig:flow} depicts the overall execution flow of our proposed setting. Given a black-box hybrid system, we learn automata as discrete abstractions of the system.
Next, we investigate the learned 
automata for critical behaviors. 
Once behaviors of interest are discovered, we use \ac{MBT} to drive the hybrid system towards these behaviors and determine its observable actions in a continuous domain. This process results in a behavioral dataset with high coverage of the hybrid system's behavior including rare conditions. 
Finally, we train a \ac{RNN} model that generalizes the behavioral dataset. 
For evaluation, we compared four different testing approaches, by generating datasets via testing, learning \ac{RNN} models from the data and computing various performance measures for detecting critical behaviors in unforeseen situations.
 Experimental results show that RNNs learned with data generated via \ac{MBT} achieved significantly better performance compared to models learned from randomly generated data. In particular, the classification error is reduced by a factor of five and a similar F1-score is accomplished with up to three orders of magnitude fewer training samples.

\begin{figure}[tb]
  \centering
  \includegraphics[width=.7\textwidth]{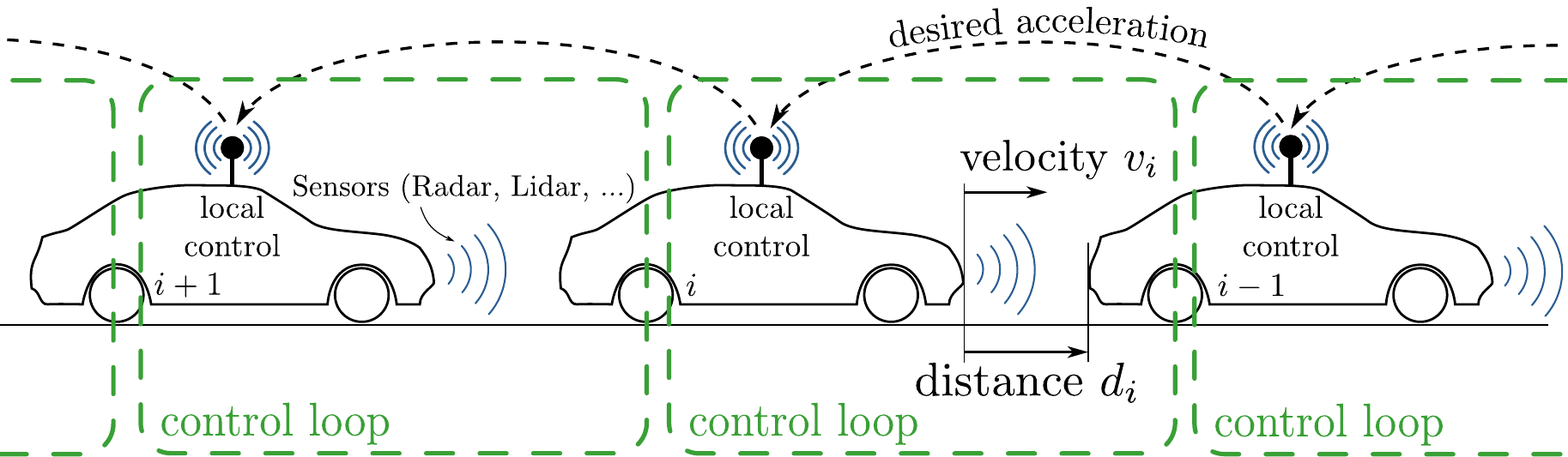}
  \caption[Platooning as distributed control scenario.]{Platooning as distributed control scenario. Adapted from a figure in~\cite{Dolk2017}.}\label{fig:platooning}
   \vspace{-0.5cm}
\end{figure}

\noindent
\textbf{Motivating Example.}
Throughout the paper we illustrate our approach utilizing a platooning scenario, implemented in a testbed in the Automated Driving Lab at Graz University of Technology (see also \url{https://www.tugraz.at/institute/irt/research/automated-driving-lab/}). 
Platooning of vehicles is a complex distributed control scenario, see~\cref{fig:platooning}. Local control algorithms of each participant are responsible for reliable velocity and distance control. 
The vehicles continuously sense their environments, e.g. the distance to the vehicle ahead and may use discrete, i.e. event triggered, communication to communicate desired accelerations along the platoon~\cite{Dolk2017}. 
Besides individual vehicle stability, the most crucial goal in controller design is to guarantee so-called string stability of the platoon~\cite{Ploeg2013}. 
This stability concept basically demands that errors in position or velocity do not propagate along the vehicle string which otherwise might cause accidents or traffic jams upstream.
Controllers for different platooning scenarios and spacing policies are available, \eg constant time headway spacing ~\cite{Ploeg2013} or constant distance spacing with or without communication~\cite{Rupp2017}.  

\me
{
Available controller designs are legitimated by rigorous mathematical stability proofs as an important theoretical foundation. In real applications it is often hard to fulfill every single modeling assumption of the underlying proofs, \eg perfect sensing or communication. However, these additional uncertainties can often be captured in fine-grained simulation models. This motivates \ac{MBT} of vehicle platooning control algorithms by the approach presented in this paper. Also, the learned behavior model can be used to detect undesired behavior during run-time. In~\cite{Dolk2017}, a hybrid system formulation of a platooning scenario is presented based on control theoretic considerations. In this contribution we aim to determine targeted behavior of such models with as few assumptions as possible by combining \ac{MBT} and machine learning. As a first step, we consider two vehicles of the platoon, the leader and its first follower, in this paper, but the general approach can be extended to more vehicles. 
}


\begin{full}
This work is an extended preprint of the conference paper ``Learning a Behavior Model of Hybrid Systems Through Combining Model-Based Testing and Machine Learning'' accepted for presentation at IFIP-ICTSS 2019, the 
31\textsuperscript{st} International Conference on Testing Software and Systems in Paris, France.
\end{full}

\me
{
\noindent
\textbf{Outline.} This paper has the following structure.
\Cref{sec:preliminaries} summarizes automata learning and \ac{MBT}.
\Cref {sec:method} explains how to learn an automaton from a black-box hybrid system, then use it to target interesting behavior of the hybrid system such that we create a behavioral dataset that can be used for machine learning purposes.
\Cref{sec:results} discusses the results gained by applying our approach to a real-world platooning scenario.
\Cref{sec:related} covers related work.
\Cref{sec:conclusion} concludes and discusses future research directions.
}

%% file: Figures/workflow.tex


\begin{tikzpicture}[transform shape,font=\scriptsize, node distance=9mm]
    \acm{\Large}
    \node [blackbox,minimum height=0.8cm, text width=1.5cm] (hybrid_system)
        at (0,0) 
        {\textbf{Hybrid System}};
    \node [whitebox,minimum height=0.8cm,  below= of hybrid_system, text width=1.5cm] (abstract_automata)
        {Finite State Model};
    \node [files,minimum height=0.8cm,  below= of abstract_automata] (tests) {Test Cases};
    \node [blackbox,minimum height=0.8cm,  text width=1.5cm, right=10mm of hybrid_system] (hybrid_system_copy)
        {\textbf{Hybrid System}};
    \node [database, database radius=5mm,database segment height=1.5mm,
            below= 0.5 of hybrid_system_copy] (dataset) {};
    \node [below=0 of dataset] (dataset_label) {\shortstack{Behavioral\\Dataset}};
    \node [whitebox,minimum height=0.8cm,text width=1.5cm] at(hybrid_system_copy|-tests) (model) {Behavior Model};
    
    \path [line] (hybrid_system) to node[left] {\shortstack{Automata\\Learning}} (abstract_automata);
    \path [line] (abstract_automata) -- node[above left = -0.25 and 0] 
        {\shortstack{Test-Case\\Generation}} (tests);
    \path [line] (tests) to ++(1.3,0) |- node[sloped, above, near start] 
        {Testing} (hybrid_system_copy);
    \path [line] (hybrid_system_copy) -- node[left=-0.05cm] {\shortstack{Logging}} (dataset);
    \path [line] (dataset_label) -- node[left=-0.05cm] {\shortstack{Machine\\Learning}} (model);
    \node[above=0.3cm of hybrid_system](mbt_label){\shortstack{Model-Based Test-Case\\Generation}};
    \node[above=0.3cm of hybrid_system_copy,text width=](b_model_label){\shortstack{Behavior Model\\Generation}};
    \draw [-, color=black!50, dotted, ultra thick] ($(mbt_label.north)!0.55!(b_model_label.north)$) -- ($(tests.south)!0.55!(model.south)$);
    
\end{tikzpicture}


%% file: Sections/preliminaries.tex
\section{Preliminaries}\label{sec:preliminaries}

\me
{
\begin{definition}[Mealy Machine]
A Mealy machine is a tuple 
$\langle \I, \O, Q,q_0,\delta,\lambda \rangle$ where $Q$ is a nonempty set of states, $q_0$ is the initial state, $\delta: Q\times \I \to Q$ is a state-transition function and $\lambda: Q\times \I \to \O$ is an output function. 
\end{definition}

\noindent We write $q\lto{i/o}q'$ if $q'=\delta(q,i)$ and $o=\lambda(q,i)$. 
We extend $\delta$ as usual to $\delta^*$ for input sequences $\pi_i$, \ie $\delta^*(q,\pi_i)$ is the state reached after executing $\pi_i$ in $q$. 
}

\me
{
\begin{definition}[Observation]
\label{def:obs}
An observation $\pi$ over input/output alphabet $\I$ and $\O$ is a pair $\langle \pi_i,\pi_o \rangle\in I^* \times O^*$ \st $\lvert\pi_i\rvert = \lvert\pi_o\rvert$. Given a Mealy machine $\M$, the set of observations of $\M$ from state $q$ denoted by $obs_\M(q)$ are
  $obs_\M(q) = \left\{ \langle \pi_i,\pi_o \rangle \in \I^* \times \O^*~\middle|~\exists\, q':q \sto{\pi_i/\pi_o} q'\right\}$,
 where $\sto{\pi_i/\pi_o}$ is the transitive and reflexive closure of the combined transition-and-output function to sequences which implies $\lvert\pi_i\rvert = \lvert\pi_o\rvert$. From this point forward, $obs_\M = obs_\M(q_0)$. Two Mealy machines $\M_1$ and $\M_2$ 
are observation equivalent, denoted $\M_1 \approx \M_2$, if $obs_{\M_1} = obs_{\M_2}$.
\end{definition}
}


\remove
{
\begin{definition}[Behavior Determinism]
\label{def:behavior}
Mealy machine $\M$ is behavior deterministic only if $\forall {\pi_o, \pi'_o \in O^*} : \langle \pi_i,\pi_o \rangle \in obs_\M \land \langle \pi_i,\pi'_o \rangle \in obs_\M \to \pi_o = \pi'_o$, that is, if $obs_\M$ contains exactly one pair $\langle \pi_i, \pi_o \rangle$ for each $\pi_i \in \I^*$.
\end{definition}
}

\subsection{Active Automata Learning}
\label{sec:preliminaries:learning}

In her semimal paper, Angluin \cite{Angluin:1987} presented \Lstar, an algorithm for
learning a \ac{DFA} accepting an unknown regular language $L$ from a \ac{MAT}.
Many other active learning algorithms also use the \ac{MAT} model~\cite{DBLP:conf/dagstuhl/HowarS16}. 
An \ac{MAT} generally needs to be able to answer two types of queries: \emph{membership} and \emph{equivalence} queries. 
In \ac{DFA} learning, the learner asks membership queries, checking inclusion of words in the language $L$. 
Once gained enough information, the learner builds a hypothesis automaton $\H$ and asks an equivalence query, checking whether $\H$ accepts exactly $L$. 
The \ac{MAT} either responds with \emph{yes}, meaning that learning was successful. Otherwise it responds with a counterexample to equivalence, \ie a word in the symmetric difference between $L$ and the language accepted by $\H$. If provided with a counterexample, the learner integrates it into its knowledge and starts a new round of learning by issuing membership queries, which is again concluded by a new equivalence query.
\Lstar is adapted to learn Mealy machines by Shahbaz and Groz~\cite{ShahbazG09}. 
The basic principle remains the same, but \emph{output queries} replace membership queries asking for outputs produced in response to input sequences. The goal in this adapted \Lstar algorithm, is to learn a Mealy machine that is observation equivalent to a black-box \ac{SUL}.

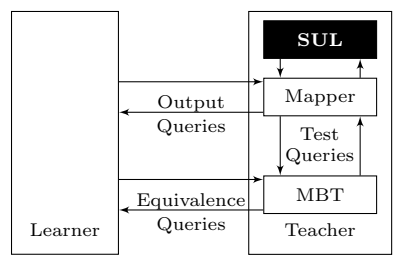
\begin{wrapfigure}[11]{t}{.44\textwidth}
  \centering
    \input{Figures/LStar_Mapper}
\vspace{-0.2cm}
  \caption[Abstract automata learning through a mapper.]{Abstract automata learning through a mapper~\cite{Vaandrager17}.
  }\label{fig:abstract_model_learning}
\end{wrapfigure}

\me
{
\paragraph*{Abstraction.}
\Lstar is only affordable for small alphabets; hence, Aarts et al.~\cite{Aarts:2012} suggested to abstract away the concrete domain of the data, by forming equivalence classes in the alphabets. This is usually done by a mapper placed in between the learner and the SUL; see \cref{fig:abstract_model_learning}.
Practically, mappers are state-full components transducing symbols back and forth between abstract and concrete alphabets using constraints defined over different ranges of concrete values. Since the input and output space of control systems is generally large or of unbounded size, we also apply abstraction by using a mapper. 

The mapper communicates with the SUL via the concrete alphabet and with the learner via the abstract alphabet.
In the setting shown in \cref{fig:abstract_model_learning}, the learner behaves like the \Lstar algorithm by Shahbaz and Groz \cite{ShahbazG09},
but the teacher answers to the queries by interacting with the SUL through the mapper.
}

\setlength{\intextsep}{0pt}%

\paragraph*{Learning and \acl{MBT}.}
Teachers are usually implemented via testing to learn models of black-box systems, 
The teacher in \cref{fig:abstract_model_learning} wraps the \ac{SUL}, uses a mapper for abstraction and includes a \acf{MBT} component.
Output queries typically reset the \ac{SUL}, execute a sequence of inputs and collect the produced outputs, \ie they perform a single test of the \ac{SUL}. Equivalence queries are often approximated via \ac{MBT}~\cite{DBLP:conf/dagstuhl/AichernigMMTT16}. 
For that, an \ac{MBT} component derives test cases (test queries) from the hypothesis model, which are executed to find discrepancies between the \ac{SUL} and the learned hypothesis, \ie to find counterexamples to equivalence. 

Various \ac{MBT} techniques have been applied in active automata learning, like the \textsc{W-Method}~\cite{Chow1978,Vasilevskii1973}, or the \textsc{partial W-Method}~\cite{Fujiware_et_al_1991}, which are also implemented in LearnLib~\cite{DBLP:conf/cav/IsbernerHS15}. These techniques attempt to prove conformance relative to some bound on
the \ac{SUL} states. 
However, these approaches require a large number of tests. Given the limited testing time available in practice, it is usually necessary to aim at ``finding counterexamples fast''~\cite{DBLP:conf/isola/HowarSM10}. Therefore, randomized testing has recently shown to be successful in the context of automata learning, such as a randomized conformance testing technique~\cite{SmeenkMVJ15} and fault-coverage-based testing~\cite{AichernigMutLearn2018}. We apply a variation of the latter, which combines transition coverage as test selection criterion with randomization. 

While active automata learning relies on \ac{MBT} to implement equivalence queries, it also enables \ac{MBT}, by learning models that serve as basis for testing~\cite{DBLP:conf/dagstuhl/AichernigMMTT16,DBLP:conf/dagstuhl/HowarS16}. Automata learning can be seen as collecting and incrementally refining information about a \ac{SUL} through testing. This process is often combined with formal verification of requirements, both at runtime and also offline using learned models. This combination has been pioneered by Peled et al.~\cite{DBLP:journals/jalc/PeledVY02} and called black-box checking. More generally, approaches that use automata learning for testing are also referred to as \ac{LBT}~\cite{DBLP:conf/dagstuhl/Meinke16}.


%% file: Figures/LStar_Mapper.tex
\begin{tikzpicture}[scale=1,font=\scriptsize, transform shape, node distance=4cm]
    \acm{\large}
    \node [] (lbl_teacher) at (0,0) {\acm{\Large} Teacher};
    \node [draw,rectangle,minimum height=0.5cm, minimum width=1.5cm, above=0/2 of lbl_teacher] (CT) {MBT};
    \node [draw,rectangle,minimum height=0.5cm, minimum width=1.5cm, above=1.3 of lbl_teacher] (mapper) {Mapper};
    \node [draw,rectangle,minimum height=0.5cm, minimum width=1.5cm, above=1/4 of mapper, fill=black, text=white] (SUL) {\bf SUL};
    \node [draw,rectangle, minimum width=1.9cm, fit= (SUL)(CT)(lbl_teacher)(mapper)] (teacher)  {};

    \node [left=2.2 of lbl_teacher] (lbl_learner) {\acm{\Large} Learner};
    \coordinate (learnernorth) at (lbl_learner|-SUL.north);
    \node [draw,rectangle, minimum width=1.2cm, fit=(lbl_learner)(learnernorth)] (learner)  {};
    
    \coordinate (mq_teacher1) at (mapper.165);
    \coordinate (mq_teacher2) at (mapper.195) {};
    \coordinate (mq_learner1) at (mq_teacher1 -| learner.east) {};
    \coordinate (mq_learner2) at at (mq_teacher2 -| learner.east);

    \node [above right=-0.25 and -0.35 of CT] (tq_ct1) {};
    \node [below right=-0.25 and -0.35 of SUL] (tq_sul1) {};
    \node [below right=-0.25 and -0.35 of mapper] (tq_mapper11) {};
    \node [above right=-0.25 and -0.35 of mapper] (tq_mapper12) {};
    \node [above left=-0.25 and -0.35 of CT] (tq_ct2) {};
    \node [below left=-0.25 and -0.35 of SUL] (tq_sul2) {};
    \node [below left=-0.25 and -0.35 of mapper] (tq_mapper21) {};
    \node [above left=-0.25 and -0.35 of mapper] (tq_mapper22) {};

    \coordinate   (eq_teacher1) at (CT.165);
    \coordinate  (eq_learner1) at (eq_teacher1-|learner.east);
    \coordinate   (eq_teacher2) at (CT.195);
    \coordinate  (eq_learner2) at (eq_teacher2-|learner.east);

    \path [line] (mq_learner1) -- (mq_teacher1);
    \path [line] (mq_teacher2) to node[below=-0.35] {\shortstack{Output\\Queries}} (mq_learner2);
    \path [line] (eq_learner1) -- (eq_teacher1);
    \path [line] (eq_teacher2) to node[below=-0.35] {\shortstack{Equivalence\\Queries}} (eq_learner2);
    \path [line] (tq_ct1) -- (tq_mapper11);
    \path [line] (tq_mapper21) to node [align=center,text width=1.2cm,right=-0.2] {Test Queries} (tq_ct2);
    \path [line] (tq_mapper12) -- (tq_sul1);
    \path [line] (tq_sul2) -- (tq_mapper22);
    
\end{tikzpicture} 

%% file: Sections/method.tex
\section{Methodology}\label{sec:method}
Our goal is to learn a behavior model capturing the targeted behavior of a hybrid \ac{SUL}. 
The model's response to a trajectory of input variables, (\eg sensor information), shall conform to  the \ac{SUL}'s response with high accuracy and precision. 
As in discrete systems, purely random generation of input trajectories is unlikely to exercise the  \ac{SUL}'s state space adequately. 
Consequently, models learned from random traces cannot accurately capture the \ac{SUL}'s behavior. 
Therefore, we propose to apply automata learning followed by \ac{MBT} to collect system traces while using a machine learning method (\ie Recurrent Neural Networks) for model learning. \Cref{fig:flow} shows a generalized version of our approach.

Our trace-generation approach does not require any knowledge, 
like random sampling, 
but may benefit from domain knowledge and specified requirements. For instance, we do not explore states any further, which already violate safety requirements. In the following, we will first discuss the testing process.
This includes interaction with the \ac{SUL}, abstraction, automata learning and test-case generation.
Then, we discuss learning a behavior model in the form of a Recurrent Neural Network with training data collected by executing tests. 

\motExample
We learn a behavior model for our platooning scenario in three steps: (1) automata learning exploring a discretized platooning control system to capture the state space structure in learned models, (2) \ac{MBT} exploring the state space of the learned model directed towards targeted behavior while collecting non-discrete system traces. In step (3), we generalize from those trace by learning a Recurrent Neural Network. 

\subsection{Testing Process}
\begin{figure*}[t]
    \centering
    \resizebox{0.9\textwidth}{!}
    {
        \input{Figures/Mapper_TCG}
    }
    \vspace{-3mm}
    \caption{Components involved in the testing process}\label{fig:tcex}
     \vspace{-6mm}
\end{figure*}
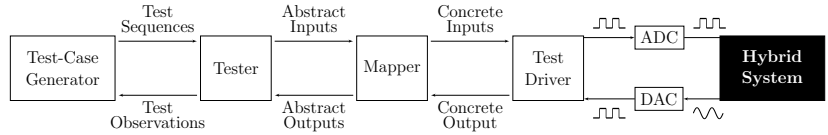
We apply various test-case generation methods, with the same underlying abstraction and execution framework. 
\Cref{fig:tcex} depicts the components implementing the testing process.

\begin{itemize}[leftmargin=*,topsep=0pt,partopsep=0pt,itemsep=0pt, parsep=0pt]
    \item {\bf Test-Case Generator:} the test-case generator creates abstract test cases. These test-cases are generated offline as sequences of abstract inputs. 
    \item {\bf Tester:} the tester takes an input sequence and passes it to the mapper. Feedback from test-case execution is forwarded to the test-case generator.
    \item {\bf Mapper:} the mapper maps each abstract input to a concrete input variable valuation and a duration, defining how long the input should be applied. Concrete output variable valuations observed during testing are mapped to abstract outputs. Each test sequence produces an abstract output sequence which is returned to the tester.
    \item {\bf Test Driver \& Hybrid System:} The test driver interacts with the hybrid system by setting input variables and sampling output variable values.
\end{itemize}

\subsubsection{System Interface and Sampling.}
\label{sec:assumptions_sampling}
We assume a system interface comprising two sets of real-valued variables: input variables $U$ and observable output variables $Y$, with $U$ further partitioned into controllable variables $U_C$ and uncontrollable, observable input variables $U_E$ affected by the environment. 
We denote all 
observable variables by $\textit{Obs} = Y \cup U_E$. Additionally, we assume the ability to \emph{reset} the \ac{SUL}, as all test runs for trace generation need to start from a unique initial state. During testing, we change the controllable variables $U_C$  and observe the evolution of variable valuations at fixed sampling intervals of length $t_s$. 

\motExample
We implemented our platooning \ac{SUL} in MathWorks Simulink\textregistered{}. The implementation actually models a platoon of remote-controlled trucks used in our testbed at the Automated Driving Lab, therefore the acceleration values and distance have been downsized.
The \ac{SUL} interface comprises: $U_C = \{ \acc \}$, $Y = \{ \dist, \vleader, \vfollow \}$, and $U_E = \{\orientation \}$.
The leader acceleration `$\acc$' is the 
single controllable input with values ranging from $\nicefrac{-1.5m}{s^2}$ to $\nicefrac{1.5m}{s^2}$, the distance
between leader and first follower is `$\dist$' and
`$\vleader$' and `$\vfollow$' are the velocities of the leader and the follower, respectively; 
finally `$\orientation$' denotes the angle between the leader and the x-axis in a fixed coordinate systems given in radians, \ie it represents the orientation of the leader that changes while driving around curves. We sampled values of these variables at fixed discrete time steps, which are $t_s = 250$ milliseconds apart. 

\subsubsection{Abstraction.}
We discretize variable valuations for testing via the mapper.
With that, we effectively abstract the hybrid system such that a Mealy machine over an abstract alphabet can model it. 
Each abstract input is mapped to a concrete valuation for $U_C$ and a duration specifying
how long the valuation shall be applied, thus $U_C$ only takes values from a finite set. \begin{conference}
As abstract inputs are mapped to uniquely defined concrete inputs, this form of abstraction does not introduce non-determinism. 
\end{conference}
In contrast, values of observable variables $\textit{Obs}$ are not restricted to a finite set.
Therefore, we group concrete valuations of  $\textit{Obs}$ and 
assign an abstract output label to each group. 

The mapper also defines a set of labels $\Violation$ containing abstract outputs that signal
violations of assumptions or safety requirements. 
In the abstraction to a Mealy machine, these outputs lead to trap states from which the model does not transit away. 
Such a policy prunes the abstract state space.

A mapper has five components: (1) an abstract input alphabet $\AI$, (2) a corresponding concretization function $\conc$, (3) an abstraction function $\abs$ mapping concrete output values to (4) an abstract output alphabet $\AO$, and (5) the set $\Violation$. During testing, it performs the following two actions: 
\begin{itemize}[leftmargin=*,topsep=0pt,partopsep=0pt,itemsep=0pt, parsep=0pt]
    \item \textbf{Input Concretization:}  the mapper maps an abstract symbol $\ai \in \AI$ 
    to a pair $\conc(\ai) = (\nu, d)$, where $\nu$ is a valuation of $U_C$ and $d\in \mathbb{N}$ defining time steps, for how long $U_C$ is set according to $\nu$.
    This pair is passed to the test driver.
    \item \textbf{Output Abstraction:} the mapper receives concrete valuations $\nu$ of $\textit{Obs}$ from the test driver and maps them to an abstract output symbol $\ao = \abs(\nu)$ in $\AO$ that is passed to the tester. 
    If $\ao \in \Violation$, then the mapper stores $\ao$ in its state and maps all subsequent concrete outputs to $\ao$ until it is reset. 
\end{itemize}

The mapper state needs to be reset before every test-case execution. 
Repeating the same symbol $\ao \in \Violation$, if we have seen it once, 
creates trap states to prune the abstract state space. 
Furthermore, the implementation of the mapper contains a cache, returning abstract output sequences without \ac{SUL} interaction.

\motExample
\label{ex:plat:abst}
We tested the \ac{SUL} with six abstract inputs $\AI$: \textit{fast-acc}, \textit{slow-acc}, 
\textit{const}, $\wait$, \textit{brake} and \textit{hard-brake}, concretized by $\conc(\textit{fast-acc}) = (\acc \mapsto 1.5 m/s^2,2)$, $\conc(\textit{slow-acc}) = (\acc \mapsto 0.7 m/s^2,2)$, $\conc(\textit{const}) = (\acc \mapsto 0,2)$, $\conc(\wait) = (\acc \mapsto 0,8)$, $\conc(\textit{brake}) = (\acc \mapsto -0.7 m/s^2,2)$, and $\conc(\textit{hard-brake}) = (\acc \mapsto -1.5 m/s^2,2)$.
Thus, each input takes two time steps, except for $\wait$, which represents prolonged driving at constant speed.

The output abstraction depends on the distance $\dist$ and the leader velocity $\vleader$. If $\vleader$ is negative, we map to the abstract output \textit{reverse}. Otherwise, we partition $\dist$ into $7$ ranges with one abstract output per range, \eg the range $(-\infty,0.43 m)$ (length of a remote-controlled truck) is mapped to \textit{crash}. 
We assume that platoons do not drive in reverse. Therefore, we include \textit{reverse} in $\Violation$, such that once we observe \textit{reverse}, we ignore the subsequent behavior. We also added \textit{crash} to $\Violation$, as we are only interested in the behavior leading to a crash. 


\subsubsection{Test-Case Execution.}
The concrete test execution is implemented by a test driver. It basically generates step-function-shaped inputs signals for input variables and samples output variable values. 
For each concrete input $(\nu_j,d_j)$ applied at time $t_j$ (starting at $t_1=0 \textit{ms}$), the 
test driver sets $U_C$ according to $\nu_j$ for $d_j\cdot t_s$ milliseconds and samples the values $\nu_j'$ of observable variables $Y \cup U_E$ at time $t_j + d_j \cdot t_s - \nicefrac{t_s}{2}$. It then proceeds to time $t_{j+1} = t_j + d_j \cdot t_s$ to perform the next input if there is any. 
In that way, the test driver creates a sequence of sampled output variable values $\nu'_{j}$, one for each concrete input. This sequence is passed to the mapper for output abstraction. 

\subsubsection{Viewing Hybrid Systems as Mealy Machines.}
Our test-case execution samples exactly one output value for each input, $\nicefrac{t_s}{2}$ milliseconds before the next input, which ensures that there is an output for each input, such that input and output sequences have the same length. Given an abstract input sequence $\pi_i$ our test-case execution produces an output sequence $\pi_o$ of the same length. In slight abuse of notation, we denote this relationship by $\lambda_h(\pi_i) = \pi_o$. Hence, we view the hybrid system under test on an abstract level as a Mealy machine $\mathcal{H}_m$ with $obs_{\mathcal{H}_m} = \{\langle \pi_i, \lambda_h(\pi_i) \rangle |\pi_i \in \AI^*\}$.

\subsubsection{Learning Automata of Motivating Example.}
\label{sec:method:autlearn}
 We applied the active automata learning algorithm by Kearns and Vazirani (KV)~\cite{kearns_vazirani_1994}, implemented by LearnLib~\cite{DBLP:conf/cav/IsbernerHS15}, in combination with the  \emph{transition-coverage} testing strategy described in previous work~\cite{AichernigMutLearn2018}. We have chosen the KV algorithm, as it requires fewer output queries to generate a new hypothesis model than, \eg \Lstar~\cite{Angluin:1987}, such that more equivalence queries are performed. As a result, we can guide testing during equivalence queries more often. The \ac{TCBT} strategy is discussed below in \cref{sec:method:tcg}.

Here, our goal 
is not to learn an accurate model, but to explore the \ac{SUL}'s state space systematically through automata learning. The learned hypothesis models basically keep track of what has already been tested. Automata learning operates in rounds, alternating between series of output queries and equivalence queries. We stop this process once we performed the maximum number of tests $\Nautlearn$, which includes both output queries and test queries implementing equivalence queries. Due to the large state space of the analyzed platooning \ac{SUL}, it was infeasable to learn a complete model, hence we stopped learning when reaching the bound $\Nautlearn$, even though further tests could have revealed discrepancies.

\motExample
The learned automata also provided insights into the behavior of the platooning \ac{SUL}. A manual analysis revealed that collisions are more likely to occur, if trucks drive at constant speed for several time steps. Since we aimed at testing and analyzing the \ac{SUL} with respect to dangerous situations, we created the additional abstract $\wait$ input, which initially was not part of the set of abstract inputs. 

During active automata learning we executed approximately $\Nautlearn 260000$ 
concrete tests on the platooning \ac{SUL} in $841$ learning rounds, producing $2841$ collisions. 
In the last round, we generated a hypothesis Mealy machine with $6011$ states that we use for model-based testing. 
\begin{conference}
Generally, $\Nautlearn$ should be chosen as large as possible given the available time budget for testing, as a larger $\Nautlearn$ leads to more accurate abstract models. 
\end{conference}

\subsubsection{Test-Case Generation.}
\label{sec:method:tcg}
In the following, we describe random test-case generation for Mealy machines, which serves as a baseline. Then, we describe three different approaches to model-based test-case generation. Note that our testing goal is to explore the system's state space and to generate system traces with high coverage, with the intention of learning a neural network. Therefore, we generate a fixed number of test cases $\Ntrain$ and do not impose conditions on outputs other than those defined by the set $\Violation$ in the mapper. 
\remove{
First, we describe them briefly and then, we discuss each in more depth.
\begin{itemize}[leftmargin=*]
    \item \textbf{Random}: random test-case generation generates sequences of randomly chosen inputs with random length. 
    \item \textbf{Learning-based}: this strategy uses active automata learning as its main test-case selection criterion, \ie equivalence and output queries generate test cases. Equivalence queries are implemented using the Transition-Coverage strategy below.
    \item \textbf{Transition-Coverage}: this strategy takes an abstract Mealy machine as input and generates tests through random walks in the Mealy machine.
    \item \textbf{Output-Directed/Crash-Directed}: this strategy takes an abstract output `$\ao$' and a learned abstract Mealy machine as input and performs walks directed towards `$\ao$'.
    In the context of platooning, we generated tests directed towards collisions, i.e. the \crash~output. 
\end{itemize}
}

\paragraph{Random Testing.} Our random testing strategy generates input sequences with a length chosen uniformly at random between $1$ and the maximum length $l_\mathrm{max}$. Inputs in the sequence are also chosen uniformly at random from $\AI$. 
\paragraph{\acl{LBT}.}
The \ac{LBT} strategy 
performs automata learning as described in \cref{sec:method:autlearn}. It produces exactly those tests executed during automata learning and therefore sets $\Nautlearn$ to $\Ntrain$. While this strategy systematically explores the abstract state space of the \ac{SUL}, it also generates very simple tests during the early rounds of learning, which are not helpful for learning a behavior model in \cref{sec:method:behav_model}.

\paragraph{\acl{TCBT}.}
The \acf{TCBT} strategy uses a learned model of the \ac{SUL} as basis. Basically, we learn a model, fix that model and then generate $\Ntrain$ test sequences with the \emph{transition-coverage} testing strategy discussed in~\cite{AichernigMutLearn2018}. We use it, as it performed well in automata learning and it scales to large automata. The intuition behind it is that the combination of variability through randomization and coverage-guided testing is well-suited in a black-box setting as in automata learning. 

Test-case generation from a Mealy machine $\mathcal{M}$  with this strategy is split into two phases, a generation phase and a selection phase. The generation phase generates a large number of tests by performing random walks through $\mathcal{M}$. In the selection phase, $n$ tests are selected to optimize the coverage of the transitions of $\mathcal{M}$. Since the $n$ required to cover all transitions may be much lower than $\Ntrain$, we performed several rounds, alternating between generation and selection until we selected and executed $\Ntrain$ test cases.  

\paragraph{Output-Directed Testing.}
Our Output-Directed Testing strategy also combines random walks with coverage-guided testing, but aims at covering a given abstract output `\textit{label}'. Therefore, it is based on a learned Mealy machine 
of the \ac{SUL}. A set consisting of $\Ntrain$ tests is generated by \cref{alg:output-directed-testing}. All tests consist of a random `\textit{prefix}' that leads to a random source state $q_r$, an `\textit{interfix}' leading to a randomly chosen destination state $q'_r$ and a `\textit{suffix}' from $q'_r$ to the `\textit{label}'. 
\begin{conference}
The suffix explicitly targets a specific output, the  interfix aims to increase the coverall \ac{SUL} coverage and the random prefix introduces variability.   
\end{conference}

\begin{algorithm}[t]
\caption{Output-Directed test-case generator}\label{alg:output-directed-testing}
\begin{algorithmic}[1]
\Require $\M=\langle \AI, \AO, Q,q_0,\delta,\lambda \rangle, \textit{label} \in \AO, \Ntrain$
\Ensure \text{TestCases} : a set of test cases directed to `$\textit{label} \in \AO$'
    \State $\text{TestCases} \gets \emptyset$
   \While{$|\text{TestCases}| < \Ntrain$}
    \State $\textit{rand-len}\gets \textsc{RandomInteger}$
    \State $\textit{prefix} \gets \textsc{RandomSequence}(\AI, \textit{rand-len})$
    \State $q_r \gets \delta^*(q_0, \textit{prefix})$
    \State $q'_r \gets \textsc{RandomState}(Q)$
    \State $\textit{interfix} \gets \textsc{PathToState}(q_r, q'_r)$ \Comment{input sequence to $q_r'$}
    \If{$\textit{interfix} \neq \bot$}
        \State $\textit{suffix}\gets \textsc{PathToLabel}(q_r, \textit{label})$ \Comment{input sequence to \textit{label}}
        \If{$\textit{suffix} \neq \bot$}
        \State $\text{TestCases} \gets \text{TestCases} \cup \{\textit{prefix} \cdot 
        \textit{interfix} \cdot \textit{suffix}$\}
        \EndIf
    \EndIf
    \EndWhile
\end{algorithmic}
\end{algorithm}

\motExample
In our platooning scenario, we aim at covering behavior relevant to collisions, 
thus we generally set $\textit{label} = \crash$ and refer to the corresponding test strategy also as Crash-Directed Testing.

%% file: Figures/Mapper_TCG.tex


\begin{tikzpicture}[scale=1,transform shape, node distance=2.5cm]
    \acm{\Large}
    \Large
    \node [whitebox,minimum height=1.9cm,text width=2.75cm] (tcg) at (0,0) {Test-Case Generator};
    \node [whitebox,minimum height=1.9cm, right= of tcg] (tester) {Tester};
    \node [whitebox,minimum height=1.9cm, right= of tester] (mapper) {Mapper};
    \node [whitebox,minimum height=1.9cm, right= of mapper] (driver) {Test Driver};
    \node [draw, minimum width=1cm, above right= -2/5 and 1.5cm of driver] (adc) {ADC};
    \node [draw, minimum width=1cm, below right= -2/5 and 1.5cm of driver] (dac) {DAC};
    \node [blackbox,text width=2.75cm, minimum height=1.9cm, right=4cm of driver] (sul) {\bf Hybrid System};
    
    \node [above right= -1/3 and -1/4 of tcg] (tcg_tester_out) {};
    \node [above left= -1/3 and -1/4 of tester] (tester_tcg_in) {};
    \node [below left= -1/3 and -1/4 of tester] (tester_tcg_out) {};
    \node [below right= -1/3 and -1/4 of tcg] (tcg_tester_in) {};
    \path [line] (tcg_tester_out) to node[above] {\shortstack{Test\\Sequences}} (tester_tcg_in);
    \path [line] (tester_tcg_out) to node[below] {\shortstack{Test\\Observations}} (tcg_tester_in);
    
    \node [above right= -1/3 and -1/4 of tester] (tester_mapper_out) {};
    \node [above left= -1/3 and -1/4 of mapper] (mapper_tester_in) {};
    \node [below right= -1/3 and -1/4 of tester] (tester_mapper_in) {};
    \node [below left= -1/3 and -1/4 of mapper] (mapper_tester_out) {};
    \path [line] (tester_mapper_out) to node[above] {\shortstack{Abstract\\Inputs}} 
    (mapper_tester_in);
    \path [line] (mapper_tester_out) to node[below] 
    {\shortstack{Abstract\\Outputs}}
    (tester_mapper_in);
    
    \node [above right= -1/3 and -1/4 of mapper] (mapper_driver_out) {};
    \node [above left= -1/3 and -1/4 of driver] (driver_mapper_in) {};
    \node [below right= -1/3 and -1/4 of mapper] (mapper_driver_in) {};
    \node [below left= -1/3 and -1/4 of driver] (driver_mapper_out) {};
    \path [line] (mapper_driver_out) to node[above] 
    {\shortstack{Concrete\\Inputs}}
    (driver_mapper_in);
    \path [line] (driver_mapper_out) to node[below] {\shortstack{Concrete\\Output}} (mapper_driver_in);
    
    \node [left=1.5cm of adc] (driver_adc_out) {};
    \node [left=1.5cm of dac] (driver_dac_in) {};
    \path [line] (driver_adc_out) to node[above] 
    {\includegraphics[page=2,width=1cm,height=5mm,trim={0 2cm 0 2cm},clip]{Figures/sin_step}} (adc);
    \path [line] (dac) to node[below] 
    {\includegraphics[page=2,width=1cm,height=5mm,trim={0 2cm 0 2cm},clip]{Figures/sin_step}} (driver_dac_in);

    \node [right=1.5cm of adc] (sul_adc_in) {};
    \node [right=1.5cm of dac] (sul_dac_out) {};
    \path [line] (adc) to node[above] 
    {\includegraphics[page=2,width=1cm,height=5mm,trim={0 2cm 0 2cm},clip]{Figures/sin_step}} (sul_adc_in);
    \path [line] (sul_dac_out) to node[below] 
    {\includegraphics[page=1,width=1cm,height=5mm,trim={0 5mm 0 5mm},clip]{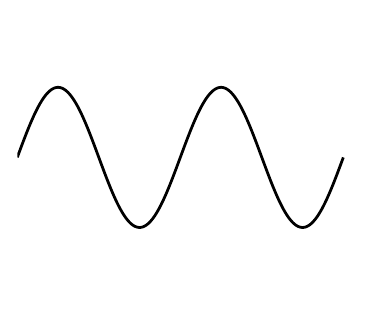}} (dac);
\end{tikzpicture} 


%% file: Sections/method_rnn_full.tex
\subsection{Learning a Behavior Model}
\label{sec:method:behav_model}
\subsubsection{Recurrent Neural Networks} \label{sec:rnns}
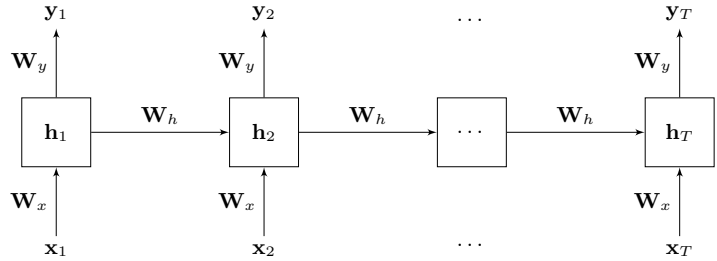
\begin{figure}
    \centering
    \resizebox{0.8\columnwidth}{!}
    {
        \input{Figures/LSTM}
    }
    \caption{Recurrent neural network. The output $\mathbf{y}_i$ of the RNN at time step $i$ does not only depend on the input $\mathbf{x}_i$ at time step $i$, but also on the accumulated knowledge in the hidden state vector $\mathbf{h}_{i-1}$ at the previous time step $i-1$.}
    \label{fig:lstm}
\end{figure}

In our scenario, we are given length $T$ sequences of vectors $\mathbf{X}=(\mathbf{x}_1, \ldots, \mathbf{x}_T)$ with $\mathbf{x}_i \in \mathbb{R}^{d_x}$ representing the inputs to the hybrid system, and the task is to predict corresponding length $T$ sequences of target vectors $\mathbf{T}=(\mathbf{t}_1, \ldots, \mathbf{t}_T)$ with $\mathbf{t}_i \in \mathbb{R}^{d_y}$ representing the outputs of the hybrid system. Recurrent neural networks (RNNs) are a popular choice for modelling these kinds of problems. In the simplest case, a plain RNN with one hidden layer of $d_h$ neurons consists of three sets of weight matrices $\mathbf{W}_x \in \mathbb{R}^{d_{h} \times d_{x}}$, $\mathbf{W}_h \in \mathbb{R}^{d_h \times d_h}$, and $\mathbf{W}_y \in \mathbb{R}^{d_y \times d_h}$ and two bias vectors $\mathbf{b}_h \in \mathbb{R}^{d_h}$ and $\mathbf{b}_y  \in \mathbb{R}^{d_y}$ which we collectively denote as $\mathbf{\Theta} = (\mathbf{W}_x, \mathbf{W}_h, \mathbf{W}_y, \mathbf{b}_h, \mathbf{b}_y)$. The RNN defines a function $\mathbf{Y}(\mathbf{X})=(\mathbf{y}_1, \dots, \mathbf{y}_T)$ recursively as
\begin{equation}
    \mathbf{y}_i = \mathbf{W}_y \mathbf{h}_i + \mathbf{b}_y, \label{eq:rnn_output_1}
\end{equation}
where $\mathbf{h}_i$ denotes the hidden state vector at time $i$ computed as
\begin{equation}
    \mathbf{h}_i = \sigma(\mathbf{W}_x \mathbf{x}_i + \mathbf{W}_h \mathbf{h}_{i-1} + \mathbf{b}_h), \label{eq:rnn_output_2}
\end{equation}
where we define $\mathbf{h}_0 = \mathbf{0}$ and $\sigma$ is an arbitrary non-linear function, e.g., $\sigma(\cdot)=\tanh(\cdot)$. The RNN architecture is depicted in~\cref{fig:lstm}. Given a set of $N$ training input/output sequence pairs $\mathcal{D}=\{(\mathbf{X}_n,\mathbf{T}_n)\}_{n=1}^N$, the task of machine learning is to find suitable parameters $\mathbf{\Theta}$ such that the output sequences $\{\mathbf{Y}_n\}_{n=1}^N$ computed by the RNN for input sequences $\{\mathbf{X}_n\}_{n=1}^N$ closely match their corresponding target sequences $\{\mathbf{T}_n\}_{n=1}^N$, and, more importantly, generalize well to sequences that are not part of the training set $\mathcal{D}$, i.e., the RNN produces accurate results on unseen data.

To obtain suitable RNN parameters $\mathbf{\Theta}$, we typically minimize a loss function describing the misfit between predictions $\mathbf{Y}$ and ground truth targets $\mathbf{T}$. A common loss function for real-valued target values $\mathbf{T}$ is the mean-squared error (MSE) loss defined by
\begin{equation}
    l(\mathbf{\Theta}, \mathcal{D}) = \frac{1}{N}\sum_{n=1}^{N} \sum_{i=1}^T \| \mathbf{y}_i(\mathbf{x}_i, \mathbf{\Theta}) - \mathbf{t}_i \|^2. \label{eq:mse_loss}
\end{equation}
Minimizing \eqref{eq:mse_loss} can be achieved by gradient descent, i.e., by iteratively correcting the current parameters $\mathbf{\Theta}$ into the direction of steepest descent,
\begin{equation}
    \mathbf{\Theta} \leftarrow \mathbf{\Theta} - \eta \ \nabla_{\mathbf{\Theta}} l(\mathbf{\Theta}, \mathcal{D}), \label{eq:gradient_descent}
\end{equation}
where $\eta$ is the learning rate that determines how much the parameters $\mathbf{\Theta}$ are changed in each iteration. Since we expect the loss function to decrease in each iteration, we obtain an RNN that gradually produces more accurate predictions on the training set $\mathcal{D}$. However, computing the gradient for the entire data set is computationally prohibitive if $N$ is large. In practice, it turns out that \emph{stochastic gradients} computed more efficiently from smaller subsets of the training data, called mini-batches, are sufficient to decrease the loss function to obtain a well-performing RNN. To this end, the entire training data is split randomly into mini-batches such that several iterations of \eqref{eq:gradient_descent} are performed with randomly selected subsets from $\mathcal{D}$. After processing the entire data, i.e., a training \emph{epoch}, the training set is split differently into a random set of mini-batches to continue with the next epoch. This minimization procedure is known as stochastic gradient descent.

\motExample
In our platooning scenario, the inputs $\mathbf{x}_i \in \mathbb{R}^2$ at time step $i$ to the hybrid system comprise the input variables $U$ from \cref{sec:assumptions_sampling}, i.e., the acceleration value $\acc$ and the orientation $\orientation$ of the leader car in radians. We preprocess the orientation $\orientation$ to contain the angular difference of orientation in radians $\orientation_i - \orientation_{i-1}$ of consecutive time steps to get rid of discontinuities when these values are constrained to a fixed interval of length $2 \pi$. The outputs $\mathbf{y}_i \in \mathbb{R}^3$ at time step $i$ of the hybrid system comprise the values of observable output variables $Y$ from \cref{sec:assumptions_sampling}, \ie the velocity of the leader $\vleader$ and the first follower $\vfollow$, respectively, as well as the distance $\dist$ between the leader and the first follower.

Furthermore, we scale and shift each input and output dimension, respectively, to have zero-mean and unit variance across all $N$ training samples and $T$ time steps. This is necessary as otherwise single input dimensions whose range is larger than the range of other dimensions tend to dominate the prediction process. The same is true for the output values since here the loss function \eqref{eq:mse_loss} could be dominated by a few output dimensions whose range is large. For prediction, the outputs of the neural networks are merely scaled and shifted inversely to their original range.

RNNs as defined in \eqref{eq:rnn_output_1}--\eqref{eq:rnn_output_2} are not constrained to sequences of a fixed length. However, from a practical perspective, training fixed-length sequences is substantially more efficient as state-of-the-art machine learning frameworks rely on GPU computation that exhibits its full parallelization potential only if data is stored in homogeneous data arrays, i.e., large tensors where all data samples are of the same size. Hence, during test case generation, we perform sequence padding at the end of the sequence by setting $\acc=0$ resulting in the leader to continue driving at approximately constant speed depending on its current orientation $\orientation$ and observing the output of the hybrid system. In rare cases the generated test sequences result in awkward behavior that needed to be truncated at some time step, e.g., when the leaders velocity $\vleader$ became negative. For these sequences, we perform a padding at the beginning of the sequence by copying the initial state where all cars have zero velocity. We used this padding procedure to obtain fixed-length sequences with $T=256$.

In our experiments we use RNNs with one hidden layer of 100 neurons. However, since plain RNNs as described in~\cref{sec:rnns} are well-known to lack the ability to model long-term dependencies, we use long short-term memory (LSTM) cells for the hidden layers \cite{Hochreiter1997}. LSTM cells are neurons equipped with an internal state and a mechanism that allows to forget, overwrite, or to keep this state for several time steps to model long-term dependencies. This functionality comes at the cost of additional weight matrices that increase the number of parameters $\mathbf{\Theta}$. For details, the interested reader is referred to \cite{Hochreiter1997}.

To evaluate the quality of the generated training sequences, we train models for several values of training set sizes $\Ntrain$. 
We used ADAM \cite{Kingma2015} implemented in Keras~\cite{chollet2015keras} with a learning rate $\eta=10^{-3}$ to perform stochastic gradient descent for 500 epochs. The number of training sequences per mini-batch is set to $\min(\Ntrain/100,500)$, i.e., equally many parameter updates per epoch are performed according to \eqref{eq:gradient_descent} up to $\Ntrain=50000$ to enhance comparability. Each experiment is performed ten times using different random initial parameters $\mathbf{\Theta}$ and we report the average performance measures over these ten runs.

%% file: Figures/LSTM.tex


\begin{tikzpicture}[scale=1,transform shape, node distance=2cm]
    \acm{\Large}
    \node[box] (h1) {$\mathbf{h}_1$};
    \node[box, right= of h1] (h2) {$\mathbf{h}_2$};
    \node[box, right= of h2] (hd) {$\dots$};
    \node[box, right= of hd] (hn) {$\mathbf{h}_T$};
    \path[line] (h1) -- node[auto]{$\mathbf{W}_h$} (h2);
    \path[line] (h2) -- node[auto]{$\mathbf{W}_h$} (hd);
    \path[line] (hd) -- node[auto]{$\mathbf{W}_h$} (hn);
    
    \node[below=1cm of h1] (x1) {$\mathbf{x}_1$};
    \node[below=1cm of h2] (x2) {$\mathbf{x}_2$};
    \node[below=1cm of hd] (xd) {$\dots$};
    \node[below=1cm of hn] (xn) {$\mathbf{x}_T$};
    \path[line] (x1) -- node[auto]{$\mathbf{W}_x$} (h1);
    \path[line] (x2) -- node[auto]{$\mathbf{W}_x$} (h2);
    \path[line] (xn) -- node[auto]{$\mathbf{W}_x$} (hn);
    
    \node[above=1cm of h1] (y1) {$\mathbf{y}_1$};
    \node[above=1cm of h2] (y2) {$\mathbf{y}_2$};
    \node[above=1cm of hd] (yd) {$\dots$};
    \node[above=1cm of hn] (yn) {$\mathbf{y}_T$};
    \path[line] (h1) -- node[auto]{$\mathbf{W}_y$} (y1);
    \path[line] (h2) -- node[auto]{$\mathbf{W}_y$} (y2);
    \path[line] (hn) -- node[auto]{$\mathbf{W}_y$} (yn);
    
\end{tikzpicture} 


%% file: Sections/method_rnn_conf.tex
\subsection{Learning a Recurrent Neural Network Behavior Model}
\label{sec:method:behav_model}
\label{sec:rnns}
In our scenario, we are given length $T$ sequences of vectors $\mathbf{X}=(\mathbf{x}_1, \ldots, \mathbf{x}_T)$ with $\mathbf{x}_i \in \mathbb{R}^{d_x}$ representing the inputs to the hybrid system, and the task is to predict corresponding length $T$ sequences of target vectors $\mathbf{T}=(\mathbf{t}_1, \ldots, \mathbf{t}_T)$ with $\mathbf{t}_i \in \mathbb{R}^{d_y}$ representing the outputs of the hybrid system. Recurrent neural networks (RNNs) are a popular choice for modelling these kinds of problems.

Given a set of $N$ training input/output sequence pairs $\mathcal{D}=\{(\mathbf{X}_n,\mathbf{T}_n)\}_{n=1}^N$, the task of machine learning is to find suitable model parameters such that the output sequences $\{\mathbf{Y}_n\}_{n=1}^N$ computed by the RNN for input sequences $\{\mathbf{X}_n\}_{n=1}^N$ closely match their corresponding target sequences $\{\mathbf{T}_n\}_{n=1}^N$, and, more importantly, generalize well to sequences that are not part of the training set $\mathcal{D}$, i.e., the RNN produces accurate results on unseen data.
To obtain suitable RNN parameters, we typically minimize a loss function describing the misfit between predictions $\mathbf{Y}$ and ground truth targets $\mathbf{T}$. 
Here, we achieve this through a minimization procedure known as stochastic gradient descent which works efficiently. For details on RNN learning, we refer to the extended version of this paper~\cite{platooning_beh_model_full}.

\motExample
In our platooning scenario, the inputs $\mathbf{x}_i \in \mathbb{R}^2$ at time step $i$ to the hybrid system comprise the input variables $U$ from \cref{sec:assumptions_sampling}, \ie the acceleration value $\acc$ and the orientation $\orientation$ of the leader car in radians. We preprocess the orientation $\orientation$ and transform it to $\orientation' = \orientation_i - \orientation_{i-1}$, the angular difference of orientation in radians of consecutive time steps to get rid of discontinuities when these values are constrained to a fixed interval of length $2 \pi$. The outputs $\mathbf{y}_i \in \mathbb{R}^3$ at time step $i$ of the hybrid system comprise the values of observable output variables $Y$ from \cref{sec:assumptions_sampling}, \ie the velocity of the leader $\vleader$ and the first follower $\vfollow$, respectively, as well as the distance $\dist$ between the leader and the first follower.

Note that RNNs are not constrained to sequences of a fixed length $T$. However, training with fixed-length sequences is more efficient as it allows full parallelization through GPU computations. Hence, during test-case execution, we pad sequences at the end with concrete inputs $(\acc\mapsto 0,1)$, \ie the leader drives at constant speed at the end of every test. In rare cases the collected test data showed awkward behavior that needed to be truncated at some time step, \eg when the leader's velocity $\vleader$ became negative. We padded the affected sequences at the beginning by copying the initial state where all cars have zero velocity. We used this padding procedure to obtain fixed-length sequences with $T=256$.

In our experiments we use RNNs with one hidden layer of 100 neurons. Since plain RNNs are well-known to lack the ability to model long-term dependencies, we use long short-term memory (LSTM) cells for the hidden layers \cite{Hochreiter1997}. 
To evaluate 
the generated training sequences, we train models for several values of training set sizes $\Ntrain$. We used ADAM \cite{Kingma2015} implemented in Keras~\cite{chollet2015keras} with a learning rate $\eta=10^{-3}$ to perform stochastic gradient descent for 500 epochs. The number of training sequences per mini-batch is set to $\min(\Ntrain/100,500)$. Each experiment is performed ten times using different random initial parameters and we report the average performance measures over these ten runs.

%% file: Sections/results.tex
\section{Experimental Evaluations}\label{sec:results}
\textbf{Predicting crashes with RNNs.}
We aim to predict whether a sequence of input values results in a crash, i.e., we are dealing with a binary classification problem. 
A sequence is predicted as positive, i.e., the sequence contains a crash, if at any time step the leader-follower distance $\dist$ gets below a certain threshold.

For the evaluation, we generated validation sequences with the Output-Directed Testing strategy. 
\begin{full}
This strategy results in sequences that contain 
crashes more frequently than the other testing strategies which 
is useful to keep the class imbalance between crash and non-crash sequences in the validation set minimal. 
\end{full}
\begin{conference}
This strategy produces 
crashes more frequently than the other testing strategies which 
is useful to keep the class imbalance between crash and non-crash sequences in the validation set minimal. 
\end{conference}
We emphasize that these validation sequences do not overlap with the training sequences that were used to train the LSTM-RNN with Output-Directed Testing sequences. The validation set
\begin{full}
\footnote{This set is usually called test set in the context of machine learning, but here we adopt the term validation set to avoid confusion with model-based testing.} 
\end{full}
contains $\Nval = 86800$ sequences out of which 17092 (19.7\%) result in a crash.
\begin{conference}
This set is usually called test set in machine learning, but here we adopt the term validation set to avoid confusion with model-based testing.
\end{conference}

For the reported scores of our binary classification task we first define:
\begin{description}[leftmargin=*,topsep=0pt,partopsep=0pt,itemsep=0pt, parsep=0pt]
    \item[True Positive (TP):]~~$\#\{$~positive sequences predicted as positive~$\}$
    \item[False Positive (FP):]~~$\#\{$~negative sequences predicted as positive~$\}$
    \item[True~Negative (TN):]$\#\{$negative sequences predicted as negative$\}$
    \item[False~Negative (FN):]$\#\{$~positive sequences predicted as negative$\}$
\end{description}

We report the following four measures: (1) the classification error (CE) in \%, (2) the true positive rate (TPR), (3) the positive predictive value (PPV), and (4) the F1-score (F1). These scores are defined as
\begin{align*}
    &\mbox{CE} = \frac{\mbox{FP} + \mbox{FN}}{\Nval} \times 100 
    &&\mbox{TPR} = \frac{\mbox{TP}}{\mbox{TP} + \mbox{FN}} \\
    &\mbox{PPV} = \frac{\mbox{TP}}{\mbox{TP} + \mbox{FP}} 
    &&F1= \frac{2\mbox{TP}}{2\mbox{TP} + \mbox{FP} + \mbox{FN}}
\end{align*}
The TPR and the PPV suffer from the unfavorable property that they result in unreasonably high values if the LSTM-RNN simply classifies all sequences either as positive or negative. The F1-score is essentially the harmonic mean of the TPR and the PPV so that these odd cases are ruled out. Note that while for the CE a smaller value indicates a better performance, for the other scores TPR, PPV, and F1 a higher score, \ie closer to $1$, indicates a better performance.

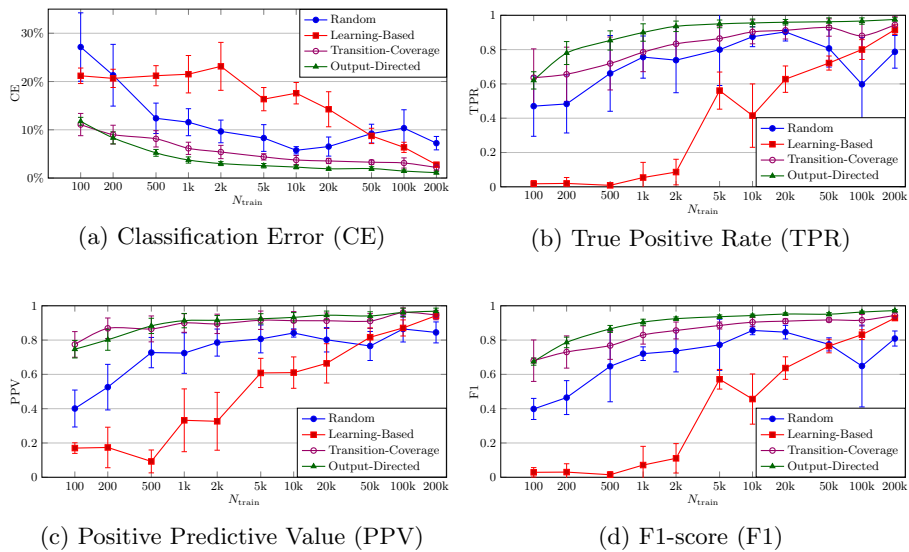
\begin{figure}[t]
\centering
\begin{subfigure}{0.5\textwidth}
    \centering
    \resizebox{\columnwidth}{!}
    {
        \input{Results/CE}
    }
    \vspace{-4mm}
    \caption{Classification Error (CE)}
    \label{fig:rnn:ce}
\end{subfigure}%
\begin{subfigure}{0.5\textwidth}
    \centering
    \resizebox{\columnwidth}{!}
    {
        \input{Results/TPR}
    }
    \vspace{-5mm}
    \caption{True Positive Rate (TPR)}
    \label{fig:rnn:tpr}
\end{subfigure}
\vfill\null\vspace{5mm}
\begin{subfigure}{0.5\textwidth}
    \centering
    \resizebox{\columnwidth}{!}
    {
        \input{Results/PPV}
    }
    \vspace{-4mm}
    \caption{Positive Predictive Value (PPV)}
    \label{fig:rnn:ppv}
\end{subfigure}%
\begin{subfigure}{0.5\textwidth}
    \centering
    \resizebox{\columnwidth}{!}
    {
        \input{Results/F1}
    }
    \vspace{-4mm}
    \caption{F1-score (F1)}
    \label{fig:rnn:f1}
\end{subfigure}
\caption{Performance measures for all testing strategies 
over changing $\Ntrain$.}
\vspace{-5mm}

\label{fig:results}
\end{figure}

The average results and the standard deviations over ten runs for these scores are shown in~\cref{fig:results}. The LSTM-RNNs trained with sequences from Random Testing and LBT perform poorly on all scores especially if the number of training sequences $\Ntrain$ is small. 
Notably, we found that sequences generated by LBT during early rounds of automata learning are short and do not contain a lot of variability, 
explaining the poor performance of LBT for low $\Ntrain$.

We can observe 
in~\cref{fig:rnn:tpr} that Random Testing and LBT perform poorly at detecting crashes when they actually occur. Especially the performance drop of LBT at $\Ntrain=10000$ and of Random Testing at $\Ntrain=100000$ indicate that additional training sequences do not necessarily improve the capability to detect crashes as crashes in these sequences still appear to be outliers. 

Training LSTM-RNNs with \ac{TCBT} and Output-Directed Testing outperforms Random Testing and LBT for all training set sizes $\Ntrain$, where the results slightly favor Output-Directed Testing. 
The advantage of \ac{TCBT} and Output-Directed Testing becomes evident when comparing the training set size $\Ntrain$ required to achieve the performance that Random Testing 
achieves using the maximum of $\Ntrain=200000$ sequences. 
The CE of Random Testing at $\Ntrain=200000$ is 7.23\% which LBT outperforms at $\Ntrain=100000$ with 6.36\%, \ac{TCBT} outperforms at $\Ntrain=1000$ with 6.16\%, and Output-Directed Testing outperforms at $\Ntrain=500$ with 5.22\%. Comparing LBT and Output-Directed Testing, Output-Directed Testing outperforms the 2.77\% CE of LBT at $\Ntrain=200000$ with only $\Ntrain=5000$ sequences to achieve a 2.55\% CE.

The F1-score is improved similarly: Random Testing with $\Ntrain=200000$ achieves 0.809, while \ac{TCBT} achieves 0.830 using only $\Ntrain=1000$ sequences, and Output-Directed Testing achieves 0.865 using only $\Ntrain=500$ sequences. Comparing LBT and Output-Directed Testing, LBT achieves 0.929 at $\Ntrain=200000$ whereas Output-Directed Testing requires only $\Ntrain=5000$ to achieve a F1-score of 0.936. In total, the sample size efficiency of \ac{TCBT} and Output-Directed Testing is two to three orders of magnitudes larger than for Random Testing and LBT.

\textbf{Evaluation of the Detected Crash Times.}
In the next experiment, we evaluate the accuracy of the crash prediction time. 
The predicted crash time is the earliest time step at which $\dist$ drops below the threshold, and the crash detection time error is the absolute difference between the ground truth crash time and the predicted crash time. Please note that the crash detection time error is only meaningful for true positive sequences.

\begin{figure}[t]
\begin{subfigure}{0.495\columnwidth}
    \resizebox{\columnwidth}{!}
    {
        \input{Results/CDFs/rand_combined}
    }
    \caption{Random}\label{fig:cdf:random}
\end{subfigure}
\begin{subfigure}{0.495\columnwidth}
    \resizebox{\columnwidth}{!}
    {
        \input{Results/CDFs/lbt_combined}
    }
    \caption{\acl{LBT}}\label{fig:cdf:lbt}
\end{subfigure}
\begin{subfigure}{0.495\columnwidth}
    \resizebox{\columnwidth}{!}
    {
        \input{Results/CDFs/tc_combined}
    }
    \caption{Transition Coverage}\label{fig:cdf:tc}
\end{subfigure}
\begin{subfigure}{0.495\columnwidth}
    \resizebox{\columnwidth}{!}
    {
        \input{Results/CDFs/out_combined}
    }
    \caption{Output-Directed}\label{fig:cdf:out}
\end{subfigure}
\caption{\ac{CDF} plots for the difference between true crash time and predicted crash time for sequences that are correctly classified as resulting in a crash.
Results are shown for all testing strategies and several training dataset sizes $\Ntrain$.}
\label{fig:cdf}
\vspace{-5mm}

\end{figure}
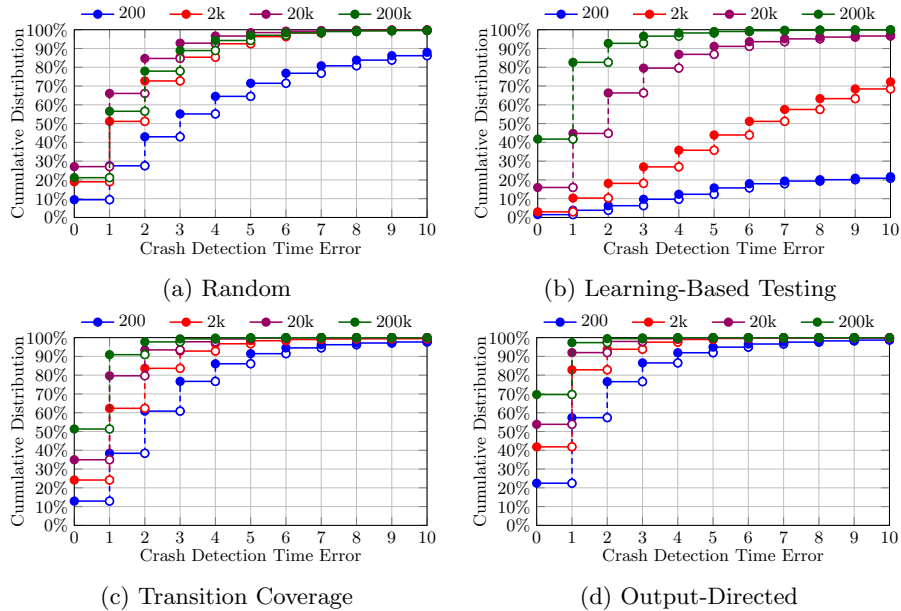

\cref{fig:cdf} shows \ac{CDF} plots describing how the crash detection time error distributes over the true positive sequences. It is desired that the \ac{CDF} exhibits a steep increase at the beginning which implies that most of the crashes are detected close to the ground truth crash time. The \ac{CDF} value at crash detection time error 0 indicates the percentage of sequences whose crash is detected without error at the correct time step.

As expected the results get better for larger training sizes $N_\text{train}$. Random Testing and LBT exhibit large errors and only relatively few sequences are classified without error. For Random Testing, less than 30\% of the crashes in the true positive sequences are classified correctly using the maximum of $N_\text{train}=200000$ sequences. On the other side, \ac{TCBT}  requires only $N_\text{train}=20000$ sequences to classify 34.9\% correctly, and Output-Directed Testing requires only $N_\text{train}=2000$ to classify 41.8\% correctly. Combining the results from~\cref{fig:cdf} with the TPR shown in~\cref{fig:rnn:tpr} strengthens the crash prediction quality even more: While \ac{TCBT} and Output-Directed Testing do not only achieve a higher TPR, they also predict the crashes more accurately. Furthermore, \ac{TCBT} and Output-Directed Testing classify 90.9\% and 97.3\% of the sequences with at most one time step error using the maximum of $\Ntrain=200000$ sequences, respectively.



%% file: Results/CE.tex
\begin{tikzpicture}
\begin{axis}[
    ymin=0, ymax=0.35, ylabel=CE,ylabel style={yshift=-5pt},
    ymajorgrids,
    xmin=50, xmax=250000,xlabel=$N_\text{train}$,xlabel style={yshift=3pt},
    yticklabel={\pgfmathparse{\tick*100}\pgfmathprintnumber{\pgfmathresult}\%},
    xtick={100,200,500,1000,2000,5000,10000,20000,50000,100000,200000},
    xticklabels={100,200,500,1k,2k,5k,10k,20k,50k,100k,200k},
    xmode=log,
    log ticks with fixed point,
    enlarge y limits=false,
    width=12cm, height=\plotheight cm, 
    legend cell align={left},
    legend style={at={(1,1)}, anchor=north east}]
    \addplot 
        plot [legend entry=Random, 
                thick, blue, 
                error bars/.cd, y dir = both, y explicit]
        table[x index=0, y index=1, y error index=2, col sep=comma]{Results/data/ce.csv};
    \addplot
        plot [legend entry=Learning-Based,
                thick, red, 
                error bars/.cd, y dir = both, y explicit]
        table[x index=0, y index=3, y error index=4, col sep=comma]{Results/data/ce.csv};
    \addplot
        plot [legend entry=Transition-Coverage, 
                mark=o, mark options={solid}, 
                smooth, thick, red!60!blue, 
                error bars/.cd, y dir = both, y explicit]
        table[x index=0, y index=5, y error index=6, col sep=comma]{Results/data/ce.csv};
    \addplot
        plot [legend entry=Output-Directed, 
                mark=triangle*, mark options={solid}, 
                smooth, thick, green!40!black, 
                error bars/.cd, y dir = both, y explicit]
        table[x index=0, y index=7, y error index=8, col sep=comma]{Results/data/ce.csv};
\end{axis}
\end{tikzpicture}

%% file: Results/TPR.tex
\begin{tikzpicture}
\begin{axis}[
    ymin=0, ymax=1, ylabel=TPR,ylabel style={yshift=-5pt},
    ymajorgrids,
    xmin=50, xmax=250000,xlabel=$N_\text{train}$,xlabel style={yshift=3pt},
    xtick={100,200,500,1000,2000,5000,10000,20000,50000,100000,200000},
    xticklabels={100,200,500,1k,2k,5k,10k,20k,50k,100k,200k},
    xmode=log,
    log ticks with fixed point,
    enlarge y limits=false,
    width=12cm, height=\plotheight cm, 
    legend cell align={left},
    legend style={at={(1,0)}, anchor=south east}]
    \addplot 
        plot [legend entry=Random, 
                thick, blue, 
                error bars/.cd, y dir = both, y explicit]
        table[x index=0, y index=1, y error index=2, col sep=comma]{Results/data/tpr.csv};
    \addplot
        plot [legend entry=Learning-Based,
                thick, red, 
                error bars/.cd, y dir = both, y explicit]
        table[x index=0, y index=3, y error index=4, col sep=comma]{Results/data/tpr.csv};
    \addplot
        plot [legend entry=Transition-Coverage, 
                mark=o, mark options={solid}, 
                smooth, thick, red!60!blue, 
                error bars/.cd, y dir = both, y explicit]
        table[x index=0, y index=5, y error index=6, col sep=comma]{Results/data/tpr.csv};
    \addplot
        plot [legend entry=Output-Directed, 
                mark=triangle*, mark options={solid}, 
                smooth, thick, green!40!black, 
                error bars/.cd, y dir = both, y explicit]
        table[x index=0, y index=7, y error index=8, col sep=comma]{Results/data/tpr.csv};
\end{axis}
\end{tikzpicture}

%% file: Results/PPV.tex
\begin{tikzpicture}
\begin{axis}[
    ymin=0, ymax=1, ylabel=PPV,ylabel style={yshift=-5pt},
    ymajorgrids,
    xmin=50, xmax=250000,xlabel=$N_\text{train}$,xlabel style={yshift=3pt},
    xtick={100,200,500,1000,2000,5000,10000,20000,50000,100000,200000},
    xticklabels={100,200,500,1k,2k,5k,10k,20k,50k,100k,200k},
    xmode=log,
    log ticks with fixed point,
    enlarge y limits=false,
    width=12cm, height=\plotheight cm, 
    legend cell align={left},
    legend style={at={(1,0)}, anchor=south east}]
    \addplot 
        plot [legend entry=Random, 
                thick, blue, 
                error bars/.cd, y dir = both, y explicit]
        table[x index=0, y index=1, y error index=2, col sep=comma]{Results/data/ppv.csv};
    \addplot
        plot [legend entry=Learning-Based,
                thick, red, 
                error bars/.cd, y dir = both, y explicit]
        table[x index=0, y index=3, y error index=4, col sep=comma]{Results/data/ppv.csv};
    \addplot
        plot [legend entry=Transition-Coverage, 
                mark=o, mark options={solid}, 
                smooth, thick, red!60!blue, 
                error bars/.cd, y dir = both, y explicit]
        table[x index=0, y index=5, y error index=6, col sep=comma]{Results/data/ppv.csv};
    \addplot
        plot [legend entry=Output-Directed, 
                mark=triangle*, mark options={solid}, 
                smooth, thick, green!40!black, 
                error bars/.cd, y dir = both, y explicit]
        table[x index=0, y index=7, y error index=8, col sep=comma]{Results/data/ppv.csv};
\end{axis}
\end{tikzpicture}

%% file: Results/F1.tex
\begin{tikzpicture}
\begin{axis}[
    ymin=0, ymax=1, ylabel=F1,ylabel style={yshift=-5pt},
    ymajorgrids,
    xmin=50, xmax=250000,xlabel=$N_\text{train}$,xlabel style={yshift=3pt},
    xtick={100,200,500,1000,2000,5000,10000,20000,50000,100000,200000},
    xticklabels={100,200,500,1k,2k,5k,10k,20k,50k,100k,200k},
    xmode=log,
    log ticks with fixed point,
    enlarge y limits=false,
    width=12cm, height=\plotheight cm, 
    legend cell align={left},
    legend style={at={(1,0)}, anchor=south east}]
    \addplot 
        plot [legend entry=Random, 
                thick, blue, 
                error bars/.cd, y dir = both, y explicit]
        table[x index=0, y index=1, y error index=2, col sep=comma]{Results/data/f1score.csv};
    \addplot
        plot [legend entry=Learning-Based,
                thick, red, 
                error bars/.cd, y dir = both, y explicit]
        table[x index=0, y index=3, y error index=4, col sep=comma]{Results/data/f1score.csv};
    \addplot
        plot [legend entry=Transition-Coverage, 
                mark=o, mark options={solid}, 
                smooth, thick, red!60!blue, 
                error bars/.cd, y dir = both, y explicit]
        table[x index=0, y index=5, y error index=6, col sep=comma]{Results/data/f1score.csv};
    \addplot
        plot [legend entry=Output-Directed, 
                mark=triangle*, mark options={solid}, 
                smooth, thick, green!40!black, 
                error bars/.cd, y dir = both, y explicit]
        table[x index=0, y index=7, y error index=8, col sep=comma]{Results/data/f1score.csv};
\end{axis}
\end{tikzpicture}

%% file: Results/CDFs/rand_combined.tex

\begin{tikzpicture}
\acm{\LARGE}
\begin{axis}[ymin=0, ymax=100, ylabel=Cumulative Distribution, 
    ylabel style={yshift=-5pt},ymajorgrids,ytick={0,10,20,...,100},
    yticklabel={\pgfmathprintnumber{\tick}\%},
    xmin=0, xmax=10, xlabel={Crash Detection Time Error}, 
    xlabel style={yshift=3pt},xmajorgrids,xtick={0,1,...,10},
    xticklabel={\pgfmathprintnumber{\tick}},
    log ticks with fixed point,
    enlarge y limits=false,
    width=8cm, height=\cdfheight cm,
    legend cell align={center},
    jump mark left,
    every axis plot/.style={thick, range={\pgfkeysvalueof{/pgfplots/xmin}}{\pgfkeysvalueof{/pgfplots/xmax}}},
    discontinuous,
    legend style={draw=none, fill=none, legend columns=4, 
        /tikz/column 2/.style={column sep=10pt},
        /tikz/column 4/.style={column sep=10pt},
        /tikz/column 6/.style={column sep=10pt}, 
        at={(0,1)}, 
        anchor=south west}]
    \addplot[legend entry=200, blue]
        table[x index=0, y index=1, col sep=comma]{Results/data/cdf_absolute_discrepancy_random.csv};
    \addplot[legend entry=2k, red]
        table[x index=0, y index=2, col sep=comma]{Results/data/cdf_absolute_discrepancy_random.csv};
    \addplot[legend entry=20k, red!60!blue]
        table[x index=0, y index=3, col sep=comma]{Results/data/cdf_absolute_discrepancy_random.csv};
    \addplot[legend entry=200k, green!40!black]
        table[x index=0, y index=4, col sep=comma]{Results/data/cdf_absolute_discrepancy_random.csv};
\end{axis}
\end{tikzpicture}

%% file: Results/CDFs/lbt_combined.tex

\begin{tikzpicture}
\acm{\LARGE}
\begin{axis}[ymin=0, ymax=100, ylabel=Cumulative Distribution, 
    ylabel style={yshift=-5pt},ymajorgrids,ytick={0,10,20,...,100},
    yticklabel={\pgfmathprintnumber{\tick}\%},
    xmin=0, xmax=10, xlabel={Crash Detection Time Error}, 
    xlabel style={yshift=3pt},xmajorgrids,xtick={0,1,...,10},
    xticklabel={\pgfmathprintnumber{\tick}},
    log ticks with fixed point,
    enlarge y limits=false,
    width=8cm, height=\cdfheight cm,
    legend cell align={center},
    jump mark left,
    every axis plot/.style={thick, range={\pgfkeysvalueof{/pgfplots/xmin}}{\pgfkeysvalueof{/pgfplots/xmax}}},
    discontinuous,
    legend style={draw=none, fill=none, legend columns=4, 
        /tikz/column 2/.style={column sep=10pt},
        /tikz/column 4/.style={column sep=10pt},
        /tikz/column 6/.style={column sep=10pt}, 
        at={(0,1)}, 
        anchor=south west}]
    \addplot[legend entry=200, blue]
        table[x index=0, y index=1, col sep=comma]{Results/data/cdf_absolute_discrepancy_learning_based_testing.csv};
    \addplot[legend entry=2k, red]
        table[x index=0, y index=2, col sep=comma]{Results/data/cdf_absolute_discrepancy_learning_based_testing.csv};
    \addplot[legend entry=20k, red!60!blue]
        table[x index=0, y index=3, col sep=comma]{Results/data/cdf_absolute_discrepancy_learning_based_testing.csv};
    \addplot[legend entry=200k, green!40!black]
        table[x index=0, y index=4, col sep=comma]{Results/data/cdf_absolute_discrepancy_learning_based_testing.csv};
\end{axis}
\end{tikzpicture}

%% file: Results/CDFs/tc_combined.tex

\begin{tikzpicture}
\acm{\LARGE}
\begin{axis}[ymin=0, ymax=100, ylabel=Cumulative Distribution, 
    ylabel style={yshift=-5pt},ymajorgrids,ytick={0,10,20,...,100},
    yticklabel={\pgfmathprintnumber{\tick}\%},
    xmin=0, xmax=10, xlabel={Crash Detection Time Error}, 
    xlabel style={yshift=3pt},xmajorgrids,xtick={0,1,...,10},
    xticklabel={\pgfmathprintnumber{\tick}},
    log ticks with fixed point,
    enlarge y limits=false,
    width=8cm, height=\cdfheight cm,
    legend cell align={center},
    jump mark left,
    every axis plot/.style={thick, range={\pgfkeysvalueof{/pgfplots/xmin}}{\pgfkeysvalueof{/pgfplots/xmax}}},
    discontinuous,
    legend style={draw=none, fill=none, legend columns=4, 
        /tikz/column 2/.style={column sep=10pt},
        /tikz/column 4/.style={column sep=10pt},
        /tikz/column 6/.style={column sep=10pt}, 
        at={(0,1)}, 
        anchor=south west}]
    \addplot[legend entry=200, blue]
        table[x index=0, y index=1, col sep=comma]{Results/data/cdf_absolute_discrepancy_transition_coverage_based_testing.csv};
    \addplot[legend entry=2k, red]
        table[x index=0, y index=2, col sep=comma]{Results/data/cdf_absolute_discrepancy_transition_coverage_based_testing.csv};
    \addplot[legend entry=20k, red!60!blue]
        table[x index=0, y index=3, col sep=comma]{Results/data/cdf_absolute_discrepancy_transition_coverage_based_testing.csv};
    \addplot[legend entry=200k, green!40!black]
        table[x index=0, y index=4, col sep=comma]{Results/data/cdf_absolute_discrepancy_transition_coverage_based_testing.csv};
\end{axis}
\end{tikzpicture}

%% file: Results/CDFs/out_combined.tex

\begin{tikzpicture}
\acm{\LARGE}
\begin{axis}[ymin=0, ymax=100, ylabel=Cumulative Distribution, 
    ylabel style={yshift=-5pt},ymajorgrids,ytick={0,10,20,...,100},
    yticklabel={\pgfmathprintnumber{\tick}\%},
    xmin=0, xmax=10, xlabel={Crash Detection Time Error}, 
    xlabel style={yshift=3pt},xmajorgrids,xtick={0,1,...,10},
    xticklabel={\pgfmathprintnumber{\tick}},
    log ticks with fixed point,
    enlarge y limits=false,
    width=8cm, height=\cdfheight  cm,
    legend cell align={center},
    jump mark left,
    every axis plot/.style={thick, range={\pgfkeysvalueof{/pgfplots/xmin}}{\pgfkeysvalueof{/pgfplots/xmax}}},
    discontinuous,
    legend style={draw=none, fill=none, legend columns=4, 
        /tikz/column 2/.style={column sep=10pt},
        /tikz/column 4/.style={column sep=10pt},
        /tikz/column 6/.style={column sep=10pt}, 
        at={(0,1)}, 
        anchor=south west}]
    \addplot[legend entry=200, blue]
        table[x index=0, y index=1, col sep=comma]{Results/data/cdf_absolute_discrepancy_output_directed_testing.csv};
    \addplot[legend entry=2k, red]
        table[x index=0, y index=2, col sep=comma]{Results/data/cdf_absolute_discrepancy_output_directed_testing.csv};
    \addplot[legend entry=20k, red!60!blue]
        table[x index=0, y index=3, col sep=comma]{Results/data/cdf_absolute_discrepancy_output_directed_testing.csv};
    \addplot[legend entry=200k, green!40!black]
        table[x index=0, y index=4, col sep=comma]{Results/data/cdf_absolute_discrepancy_output_directed_testing.csv};
\end{axis}
\end{tikzpicture}

%% file: Sections/related.tex
\section{Related Work}\label{sec:related}
\emph{Verifying Platooning Strategies.}
\begin{conference}
 Meinke~\cite{Meinke17} used \ac{LBT} to analyze vehicle platooning systems with respect to qualitative safety properties, like collisions.
 While the automata learning setup is similar to our approach, he aimed (1) to show how well a multi-core implementation of \ac{LBT} method scales, 
 and (2) how problem size and other factors affect scalability. 
Fermi et al.~\cite{FermiMMF18} applied rule inference methods to validate collision avoidance in platooning. 
More specifically, they used decision trees as classifiers for safe and unsafe platooning conditions, and they suggested three approaches to minimize the number of false negatives.
Rashid et al.~\cite{RashidSH18} modelled a generalized platoon controller formally in higher-order logic.
They proved satisfication of stability constraints in \tool{HOL Light} and showed how stability theorems can be used to develop runtime monitors.
\end{conference}
\begin{full}
\me
{
Meinke~\cite{Meinke17} used \ac{LBT} to analyze vehicle platooning systems from the perspective of qualitative safety properties, such as vehicle collisions. 
The intentions and motivations behind \cite{Meinke17} are: (1) to show how well a multi-core implementation of \ac{LBT} method scales, and (2) how problem size and other factors affect scalability. 
Similar to our approach, the author learned the platooning system by incorporating it as a \ac{SIL} in the learning phase.
Experiments showed promising scalability results. 
He has successfully generated testcases for an invariant property specifying the optimal distance between vehicles in a one--dimension platooning scenario, meaning that he used no steering model.
We believe that since testing and verification of the platooning control strategy were not the primary intentions of the author, and because the experimental results already fulfilled the primary purposes of the author, no further investigations were made to explain why the learned models did not generalize well
It is explicitly stated in \cite[p.~13]{Meinke17} that the learned models agree with the platoon control strategy at most up to 9.4\% of traces.
This indicates the learned models do not provide a good generalization of collision scenarios and thus cannot be used as to predict vehicle collisions.
}

\me
{
Fermi et al.~\cite{FermiMMF18} showed how to derive sensitivity of safety conditions in vehicle platooning strategies using rule inference methods. 
They have used the \tool{Plexe} simulator to generate a dataset of a platoon control strategy on which they defined a prediction problem by setting a supervised learning method to distinguish safe and unsafe platooning conditions. 
More specifically, they used \ac{DT} as the classifier, and they suggested three approaches to minimize the number of false negatives.
First, manual inspection of inferred rules based on the analysis in \cite{CangelosiMPBBVCV14} using two or three highest ranked features of their dataset, which is labor intensive.
Second, they used \ac{LLM} to infer the set of rules distinguishing types of platooning conditions with a safe margin knowing they are training \ac{LLM} with zero error. This approach is too conservative and does not provide a good generalization.
Finally, they introduced a semi-automatic approach based on the principle of $\kappa$--fold cross-validation. The authors divided the dataset into $\kappa$--folds and trained a model with a non--zero margins (\eg 5\%) and inferred a set of rules identifying safe platooning conditions and manually inspected false negative conditions refining the learned model. They used \ac{FPR} and \ac{FNR} for validation of the reliability of the prediction.
}

\me
{
Rashid et al.~\cite{RashidSH18} used higher-order logic to model a generalized platoon controller formally.
Their modeling formalism incorporates the physical analysis of the platoon using multivariate calculus and Laplace transform.
They also specified the platoon stability formally and then used \tool{HOL Light} to verify important stability constraints for arbitrary platoon parameters.
The verification approach results in stability theorems; therefore, it provides a more general analysis compared to both simulation-based approaches (\ie where verification only holds for the applied test cases) and automata-theoretic verification approaches (\ie modeling platoon control strategies by a discrete--time model using automata).
On the other hand, this verification approach requires manual modeling of the platoon controller; that is, it only ensures the correctness of an abstract model. 
For this reason, the authors investigated methods to translate the stability theorems resulted from the verification approach into runtime monitors to detect the violations of any stability constraints and runtime enforcement of correct strategies.
Finally, we believe since the runtime monitors are a byproduct of manually crafted models of the platoon controllers there is no prior knowledge about how well they perform in practice.
}

\end{full}

\noindent\emph{System Identification and State Estimation.}
Determining models by using input-output data is known as \textit{system identification} in the control systems community~\cite{Ljung1999}. Such models can be useful for simulation, controller design or diagnosis purposes. Recently, progress towards system identification techniques for hybrid systems based on the classical methods presented e.g. in the book of Ljung~\cite{Ljung1999} has been made, see~\cite{Vidal2016} and the references therein. In~\cite{Vidal2016}, single-input single-output models are considered. Furthermore, the contribution focuses on so-called piece-wise affine ARX models. We believe that the presented hybrid automata learning techniques could essentially contribute to this research field by relaxing some of the modeling assumptions.

If the model parameters and the switching mechanism is known, the problem reduces to a hybrid state estimation problem~\cite{Tanwani2013,Lv2018}. 
 Such estimators or observers are used in various problems, i.e. closed loop control, parameter estimation or diagnosis. 
 However, the traditional methods often assume accurate and exact models. 
 These models are mostly derived based on first principles~\cite{Lv2018}, 
 which is often not feasible in complex scenarios. This shows the advantage of our learning-based approach especially in cases without detailed model knowledge.

%% file: Sections/conclusion.tex
\section{Future Work \& Conclusion}\label{sec:conclusion}
We successfully combined abstract automata learning, \ac{MBT}, and machine learning to learn a behavior model from observations of a hybrid system. Given a black-box hybrid system, we learn an abstract automaton capturing its discretized state-space; then, we use \ac{MBT} to target a behavior of interest. This results in test suites with high coverage of the targeted behavior from which we generate a behavioral dataset. 
LSTM-RNNs are used to learn behavior models from the behavioral dataset.
 Advantages of our approach are demonstrated on a real-world case study; \ie a platooning scenario. 
Experimental evaluations show that LSTM-RNNs learned with model-based data generation achieved significantly better results compared to models learned from randomly generated data, \eg reducing the classification error by a factor of five, or achieving a relatively similar F1-score with up to three orders of magnitude fewer training samples than random testing. 
This is accomplished through systematic testing (\ie automata learning, and \ac{MBT}) of a black-box hybrid system without requiring a priori knowledge on its dynamics.

Motivated by the promising results shown in Sect.~\ref{sec:results}, we plan to carry out further case studies. For future research, we target runtime verification and runtime enforcement of safety properties for hybrid systems. 
To this end, we conjecture that a predictive behavior model enables effective runtime monitoring, 
which allows us to issue warnings or to intervene in case of likely safety violations.
\begin{conference}
Adaptations of the presented approach are also potential targets for future research, e.g. automata learning and test-based trace generation could be interleaved in an iterative process.
\end{conference}